\documentclass[preprint,12pt,authoryear]{elsarticle}

\usepackage{amssymb}
\usepackage{amsmath}
\usepackage{multirow}
\usepackage{booktabs}
\usepackage{subcaption} 
\usepackage{tikz}       
\usepackage{float}
\usepackage{makecell} 

\usepackage[colorlinks=true, citecolor=blue, linkcolor=blue, urlcolor=blue]{hyperref}
\usepackage[margin=1in]{geometry}
\usepackage[toc,page]{appendix}
\usepackage{algorithm}
\usepackage{algpseudocode}

\begin{document}

\begin{frontmatter}



\title{On Predicting Sociodemographics from Mobility Signals} 


\author[label1]{Ekin Uğurel} 

\affiliation[label1]{organization={Department of Civil and Environmental Engineering, University of Washington},
            city={Seattle},
            postcode={98105}, 
            state={WA},
            country={USA}}

\author[label1]{Cynthia Chen} 

\author[label2]{Brian H. Y. Lee}

\author[label3]{Filipe Rodrigues} 

\affiliation[label2]{organization={Puget Sound Regional Council (PSRC)},
            city = {Seattle},
            postcode = {98101},
            state = {WA},
            country = {USA}}

\affiliation[label3]{organization={Department of Technology, Management and Economics, Technical University of Denmark},
            city={Copenhagen},
            postcode={2800}, 
            state={Capital Region},
           country={Denmark}}

\begin{abstract}
Inferring sociodemographic attributes from mobility data could help transportation planners better leverage passively collected datasets, but this task remains difficult due to weak and inconsistent relationships between mobility patterns and sociodemographic traits, as well as limited generalization across contexts. We address these challenges from three angles. First, to improve predictive accuracy while retaining interpretability, we introduce a behaviorally grounded set of higher-order mobility descriptors based on directed mobility graphs. These features capture structured patterns in trip sequences, travel modes, and social co-travel, and significantly improve prediction of age, gender, income, and household structure over baselines features. Second, we introduce metrics and visual diagnostic tools that encourage evenness between model confidence and accuracy, enabling planners to quantify uncertainty. Third, to improve generalization and sample efficiency, we develop a multitask learning framework that jointly predicts multiple sociodemographic attributes from a shared representation. This approach outperforms single-task models, particularly when training data are limited or when applying models across different time periods (i.e., when the test set distribution differs from the training set). 
\end{abstract}




\end{frontmatter}


\newpage
\section{Introduction}
\label{sec1}


Over several decades, travel behavior scholarship has revolved around the interplay between \emph{who} a person is and \emph{how} they move. Household‐travel surveys (HTSs) have long provided the empirical backbone for this work by coupling rich trip diaries with respondent characteristics such as age, gender, income, and household composition. Analyses drawing on these surveys consistently show that, after accounting for the built environment, sociodemographic traits still correlate with car ownership, mode choice, trip frequency, and trip-chaining behavior \citep{Bhat1994structural, Lee2007Household, Lu1999Socio, McGuckin1999Examining, Mokhtarian2004TTB}

In the past dozen years, the ubiquity of GPS-enabled smartphones has spawned a parallel, industry-scale source of mobility evidence in passively-generated mobile data, which includes call-detail records (CDR), location-based service (LBS) pings, connected-vehicle traces, and the like \citep{Chen2016promises}. These datasets dwarf HTSs in both sample size and temporal length, are refreshed continuously, and can often be licensed at a fraction of the cost of running a tailored survey. Their content, however, is almost exclusively spatial temporal; they record where and when a device was observed but remain agnostic about \emph{who} was holding it. This missing dimension limits many distributional and behavioral analyses, including those that require understanding how travel patterns vary across population subgroups.

Despite this blind spot, public agencies have been keen on experimenting with mobile data products \citep{Ugurel2024Metropolitan}. Metropolitan planning organizations (MPOs) see potential in using them to stitch origin-destination matrices \citep{Alexander2015Origindestination, Iqbal2014Development}, site electric-vehicle chargers \citep{Yang2017EV}, and evaluate complete-street retrofits \citep{Bian2023Lessons}. Yet the lack of respondent attributes imposes two related hazards. First, representativeness: smartphone datasets systematically under sample certain population subgroups \citep{Li2024Understanding, Wang2025Exploring, Wesolowski2013impact, wu2024location}. As a result, decisions based solely on such data may reflect the travel patterns of overrepresented groups while ignoring others. Second, for many analyses MPOs would like to conduct, such as gauging the effects of new toll roads or assessing if expanded transit lines reach underserved communities, linking mobility traces to travelers’ sociodemographic profiles is required.

To unlock the full value of passively-generated traces, researchers need a way to infer or impute sociodemographic variables from the mobility patterns embedded in the devices. This issue has been studied from a variety of angles, including travel behavior \citep{Auld2015Demographic, Zhang2024Social}, data mining of destination choice \citep{Doi2021Estimation, Solomon2018Predict, Wu2019Inferring, Zhong2015You}, communication metadata \citep{Jahani2017Improving, Razavi2024Predicting}, transit smart card use \citep{Ding2019Estimating}, and information theory \citep{Zhao2022Theory}. As our goal is to enable the use of mobile data for transportation planning, we primarily focus on studies that solely use mobility behavior to achieve the aim of imputing sociodemographic attributes from mobility traces.

Investigating this sociodemographic-inference problem presents two principal obstacles. First, most of the foundational literature in travel behavior frames sociodemographics as shaping mobility, not the reverse. This perspective aligns with behavioral models in which age, income, and household responsibilities influence when, where and how people travel. As a result, conceptualizing mobility traces as predictive signals for inferring sociodemographic attributes remains a relatively underdeveloped and counterintuitive direction in the literature. Second, sociodemographic inference from behavioral signals is intrinsically difficult. While some features of travel (e.g., trip chaining or mode diversity) can correlate with variables like gender or income, the relationships are often weak, noisy, and highly context dependent. In practice, models trained on one dataset may exhibit strong predictive performance yet fail to generalize when applied to another dataset, especially across geographies with different urban forms and social norms \citep{Sheller2006New}. These challenges complicate efforts to construct transferable models and limit the extent to which empirical findings can be applied universally.

This study aims to overcome these obstacles through the following contributions. First, to confront the limited theoretical grounding for inferring demographics from mobility, we introduce a family of higher-order descriptors based on directed mobility graphs in which vertices encode activity purposes and edges connect chronologically adjacent trips. By \emph{higher-order}, we mean measures that go beyond first-order metrics (i.e., visit counts, mode frequencies) to capture relations across sequences of trips and destinations (e.g., how evenly travel is spread across activities, whether trips form loops or tours, whether they mix modes, etc.). We demonstrate that these descriptors are strongly and interpretablely associated with sociodemographic attributes and that they substantially increase the predictive power of imputation models beyond classical and spatiotemporal features (defined in Section \ref{sec4.1}).

Second, to mitigate the inherent difficulty and uncertainty in generalizing sociodemographic inference models across diverse contexts, we operationalize uncertainty quantification and model calibration for multi-class classification. Specifically, we leverage metrics that encourage models to match their prediction confidences with their accuracies. We also use visual diagnostic tools like reliability diagrams to identify calibration gaps in our experiments. Applying this protocol, we find that the proposed higher-order descriptors consistently improve out-of-sample likelihoods, but their effect on calibration is mixed: they close gaps in several settings but can yield conservative (under-confident) probabilities, possibly due to the risk of overfitting.

Third, we establish the value of a multitask (MT) learning strategy that predicts multiple sociodemographic attributes simultaneously. Under standard statistical learning assumptions, parameter sharing across related tasks reduces estimator variance and improves out of sample performance \citep{Baxter2000Model, Caruana1997Multitask}. Thus, jointly modeling targets like age, gender, income, and household size allows the network to exploit shared structure in the mapping from mobility to sociodemographic attributes, yielding a more data-efficient and robust estimator than training separate models. Moreover, MT learning can improve transferability across contexts by reducing sensitivity to distributional shift (i.e., changes in the input–output relationship between training and test data). We corroborate these claims with cross‑temporal generalization experiments in which the multi‑task variant consistently outperforms single‑task baselines under a range of sample‑size constraints.\\

\textbf{Our Contributions}:
\begin{itemize}
    \item We introduce a behaviorally grounded family of \textbf{higher-order descriptors based on mobility graphs}, in which vertices represent travel purposes and edges represent temporally ordered trips between them. When appended to classical mobility features, our feature set consistently raises the out-of-sample accuracy of sociodemographic inference (e.g., age, gender, income, etc.) across multiple experimental setups.
    \item We operationalize \textbf{uncertainty quantification and calibration methods} for mobility-based sociodemographic inference, using metrics that align predicted confidence with observed accuracy. Visual tools like reliability diagrams reveal calibration gaps and help diagnose model behavior.
    \item We demonstrate the benefits of \textbf{multi‑task learning} for transferability and data efficiency. Training a unified model to predict multiple sociodemographic attributes jointly improves sample efficiency and enhances generalization to new data compared to single‑task baselines. In both data sparse regimes and those in which the test set systematically differs from the training set, the multi‑task variant consistently outperforms single‑task counterparts, indicating enhanced robustness and practical transfer across contexts.
\end{itemize}

The rest of this paper is organized as follows: Section \ref{sec2} provides a review of relevant literature, both in the context of classical travel behavior studies and those that seek to answer the same question as ours. Section \ref{sec3} outlines the datasets we leverage, while Section \ref{sec4} details our methodological approach. Section \ref{sec5} presents our experimental design and the numerical results. We conclude with a discussion of our findings, limitations, and ideas for future work in Section \ref{sec6}.

\section{Literature Review}
\label{sec2}

In this section, we review previous efforts to infer sociodemographic backgrounds of individuals based on their mobility behavior. We distinguish between these studies and a parallel body of work that predicts demographics from mobile phone communication metadata (call records, texting patterns, app usage, etc.). Both lines of research share the premise that digital behavioral traces contain sociodemographic signals, but they differ in the type of behavior analyzed. Mobility focused studies use \textit{where people go} as the primary predictor, whereas communication focused studies use who people connect with and how they use their devices. 

Most mobility-inference studies begin by transforming raw coordinates into interpretable spatial or temporal descriptors. Spatial metrics quantify the extent and diversity of an individual’s movements: the radius of gyration \citep{Ding2019Estimating}, the heterogeneity of visited locations \citep{Wu2019Inferring}, the number of unique visited locations \citep{Wu2019Inferring, Zhong2015You}, and those that relate to the distance traversed \citep{Wu2019Inferring}. Temporal features characterize regularity and rhythm, such as the day-to-day similarity of location sequences or the distribution of departure times for commutes \citep{Ding2019Estimating}, under the hypothesis that highly routinized commuters differ demographically from, for example, students or gig-economy workers. Semantic signals add further discriminatory power. Studies extract the frequency of visits to certain points-of-interest (POIs). For example, frequent stops at beauty salons or supermarkets can proxy for gender \citep{Doi2021Estimation}, while regular, short school drop-offs can signal parental status. From a classical travel behavior lens, stop durations, tour counts, and land-use contexts also prove to carry predictive power on whether or not a person drives, their work status, and education level \citep{Auld2015Demographic}. Despite the range of mobility descriptors, it remains unclear which signals reliably predict which attributes and under what conditions. Reported associations are often ad hoc and context-specific. Most studies focus on overall accuracy using bundled features, without isolating the marginal value of specific groups (spatial, temporal, semantic) or testing their consistency across settings. To fill this gap, we systematically assess the incremental contribution and robustness of each feature family.

To interpret these descriptors, researchers have moved from statistical models to more data-driven approaches. Early work mapped those hand-crafted features into conventional statistical or rule-based models. \cite{Auld2015Demographic} combined fuzzy clustering of classical travel descriptors with decision trees and nested-logit models, while \cite{Zhong2015You} used tensor factorization to decompose location check-in patterns. More recent research has gravitated toward representation-learning. \cite{Solomon2018Predict} interpreted each day’s trajectory as a “sentence’’ and learned Word2Vec embeddings prior to classification, whereas \cite{Ding2019Estimating} employed long short-term memory networks on smart-card sequences. \cite{Xu2020SUME} modeled the city as a heterogeneous mobility network and learned embeddings by preserving both physical co-visitation and semantic similarity between users. Generally, these approaches aim to capture the temporal structure and contextual regularities of daily movement, letting the model implicitly learn what aspects of movement are informative for demographics. However, most of these models train separately for each attribute, ignoring shared structure across related targets (i.e., age and the number of children). This makes them data-hungry and less transferable across settings with distribution shifts. We address this by using a multitask learning framework that jointly predicts all attributes from a shared representation, improving sample efficiency and out-of-sample robustness.

When it comes to the output of prediction models, the uncertainty that is quantified needs to be interpreted with care. Many works simply present a confusion matrix or error rate, which is an aggregate uncertainty measure. These tell us, for example, that the model is wrong 20\% of the time overall, but not \textit{which} 20\% of cases or how to flag an uncertain individual. The tendency of earlier models to overfit their training data (as in \cite{Auld2015Demographic}) highlights the risks of relying on point predictions without accounting for variance or confidence. A few studies have taken steps toward more meaningful uncertainty estimates. For example, some inference models incorporate cross-validation accuracy into probability estimates, essentially adjusting predicted class probabilities downward to reflect the model’s known error rate \citep{Zhang2024Social}. This can prevent overconfidence in the predictions by “baking in” the chance of error. Moreover, the recent work by \cite{Zhao2022Theory} introduces a theoretical framework to estimate beforehand how separable the classes might be, given the covariance structure of mobility data. Building on these advances, we calibrate predicted probabilities using metrics that encourage honesty between confidence and accuracy. We further show that our feature set and multitask learning framework yield more reliable and discriminative uncertainty estimates compared to baselines. These steps support a shift toward uncertainty-aware inference, where decisions can reflect the model’s confidence.

\section{Datasets}
\label{sec3}
We analyze three of the four most recent waves of the Puget Sound Regional Council (PSRC) Household Travel Survey (HTS), with the exception being the 2021 wave during which travel behavior was heavily confounded by the COVID-19 pandemic. The 2017, 2019, and 2023 surveys were fielded biennially using an address-based probability sample covering the four-county central Puget Sound region (King, Kitsap, Pierce, and Snohomish). The 2021 wave uniquely included an additional opt-in online panel, which we omit here for consistency. Table \ref{tab1} highlights relevant descriptive statistics from our post-processed version of this dataset.

\begin{table}[ht]
\caption{Descriptive statistics of the PSRC HTS in the three waves used in this study (post-processing; values in parentheses denote the strata percentage associated with wave)}
\label{tab1}
\centering
\vspace{0.1in}
\begin{tabular}{c|r|r|r|r}
\textbf{Category} & \textbf{Variable} & \textbf{2017} & \textbf{2019} & \textbf{2023} \\
\hline
\multirow{4}{*}{\textbf{Wave}} 
 & Field Dates & 04/10 -- 06/15 & 03/11 -- 05/30 & 04/24 -- 05/29 \\
 & Households & 3,160 & 2,902 & 3,504 \\
 & Persons & 5,545 & 5,116 & 5,959 \\
 & Trips & 51,029 & 71,913 & 56,028 \\
\hline
\multirow{3}{*}{\textbf{Gender}}
 & Male & 2,714 (\textbf{48.9\%}) & 2,475 (\textbf{48.4\%}) & 2,597 (\textbf{43.6\%}) \\
 & Female & 2,724 (\textbf{49.1\%}) & 2,533 (\textbf{49.5\%}) & 2,854 (\textbf{47.9\%}) \\
 & Non-binary & 17 (\textbf{0.31\%}) & 23 (\textbf{0.45\%}) & 121 (\textbf{2.03\%}) \\
\hline
\multirow{6}{*}{\textbf{Age}}
 & 0--11 & 495 (\textbf{9.74\%}) & 477 (\textbf{10.1\%}) & 549 (\textbf{10.8\%}) \\
 & 12--17 & 151 (\textbf{2.97\%}) & 159 (\textbf{3.36\%}) & 197 (\textbf{3.93\%}) \\
 & 18--34 & 1,739 (\textbf{34.2\%}) & 1,514 (\textbf{31.9\%}) & 1,335 (\textbf{26.7\%}) \\
 & 35--54 & 1,562 (\textbf{30.7\%}) & 1,480 (\textbf{31.3\%}) & 1,517 (\textbf{30.3\%}) \\
 & 55--74 & 970 (\textbf{19.1\%}) & 941 (\textbf{19.9\%}) & 1,152 (\textbf{23.0\%}) \\
 & 75+ & 165 (\textbf{3.25\%}) & 163 (\textbf{3.44\%}) & 267 (\textbf{5.33\%}) \\
\hline
\multirow{5}{*}{\textbf{HH Income}}
 & Under \$25,000 & 410 (\textbf{8.07\%}) & 314 (\textbf{6.63\%}) & 354 (\textbf{7.07\%}) \\
 & \$25k--49,999 & 642 (\textbf{12.6\%}) & 588 (\textbf{12.4\%}) & 533 (\textbf{10.6\%}) \\
 & \$50k--74,999 & 744 (\textbf{14.6\%}) & 715 (\textbf{15.1\%}) & 646 (\textbf{12.9\%}) \\
 & \$75k--99,999 & 728 (\textbf{14.3\%}) & 657 (\textbf{13.9\%}) & 531 (\textbf{10.6\%}) \\
 & \$100k+ & 2,558 (\textbf{50.3\%}) & 2,460 (\textbf{51.9\%}) & 2,943 (\textbf{58.8\%}) \\
\hline
\multirow{4}{*}{\textbf{Children in HH}}
 & 0 & 3,538 (\textbf{69.6\%}) & 3,297 (\textbf{69.6\%}) & 3,371 (\textbf{67.3\%}) \\
 & 1 & 702 (\textbf{13.8\%}) & 586 (\textbf{12.4\%}) & 604 (\textbf{12.1\%}) \\
 & 2 & 712 (\textbf{14.0\%}) & 580 (\textbf{12.2\%}) & 777 (\textbf{15.1\%}) \\
 & 3+ & 130 (\textbf{2.57\%}) & 250 (\textbf{5.28\%}) & 277 (\textbf{5.53\%}) \\
\end{tabular}
\end{table}

During data cleaning we applied a set of exclusion rules to ensure that only complete and internally consistent trip records entered the analysis file. First, any trip lacking a valid origin or destination purpose code was deleted, because purpose is central to several of our behavioral indicators. We likewise removed trips for which the travel‐mode field was blank; mode choice is a key outcome variable and cannot be reliably imputed when entirely missing. Third, observations without usable spatial information—specifically, trips whose destination coordinate could not be geocoded or whose reported distance was zero—were discarded, as they preclude computation of distance-based measures. Finally, we excluded the small number of records reporting negative travel durations, which are symptomatic of data-entry or time-stamp errors. These filters leave a sample of trips with complete purpose, mode, spatial and temporal attributes suitable for subsequent modeling.

Although our motivation is to enable sociodemographic inference from passively-collected mobile data, we conduct this study using HTS data due to two key advantages. First, it provides ground-truth sociodemographic labels, which are essential for supervised learning and evaluation. Second, because HTS is fielded biannually, it enables testing under distribution shift by evaluating models across different survey waves. While our features are derived from structured trip diaries, recent advances in imputation algorithms now make it feasible to extract similar semantic information from raw GPS traces \citep{Gao2024Activity, Merikhipour2024Transportation}. Thus, the methods developed here can largely be applied to passive data once suitably enriched.

\section{Methodology}
\label{sec4}
Section \ref{sec4} details our methodology. \ref{sec4.1} formalizes the mobility‑graph representation and defines the activity‑ and trip‑level covariates used in prediction. In \ref{sec4.2}, we outline definitions and metrics that allow us to operationalize uncertainty quantification in our context. Finally, \ref{sec4.3} describes the theory behind the MT approach as well as the specific neural architecture we leverage.

\subsection{Characterizing travel behavior with mobility graphs}
\label{sec4.1}
Daily activity chains are encoded as directed graphs $G = (V, E)$, where vertices $V$ represent unique destination purposes (e.g., home, gym, school), and edges $E$ represent temporally ordered trips between them. From this representation, we extract two layers of information: (1) \textit{activity (node) features}, which describe how frequently, evenly, and in what sequence specific activities are pursued; and (2) \textit{trip (edge) features}, which summarize mode diversity (i.e., tendency to use different travel modes), co-travel composition (i.e., share of trips taken alone vs.\ with others), and daily travel motifs (i.e., minimal, recurring subgraphs that capture the daily structure of trip sequences). We give precise definitions for metrics to quantify the above in Section~\ref{sec4.1.2} (see Figures 1--3).

At its simplest, the fraction of one’s trips dedicated to specific activities is a classical indicator of who they may be \citep{Lu1999Socio}. Persons with children spend more time at schools, while older individuals spend more time shopping. Similarly, the fraction of trips taken with modes can be indicative of sociodemographic background. In the Seattle context, those with higher incomes tend to drive more, while men tend to bike more than women (we detail more correlations in Section \ref{sec5.1}). Though this type of knowledge is not readily available from raw GPS data, there are now imputation algorithms that can infer multiple class trip purposes and modes with F1 scores up to 76\% \citep{Gao2024Activity} and 92\% \citep{Merikhipour2024Transportation}, respectively.

\begin{figure}[!]
    \centering
    \includegraphics[width=\linewidth]{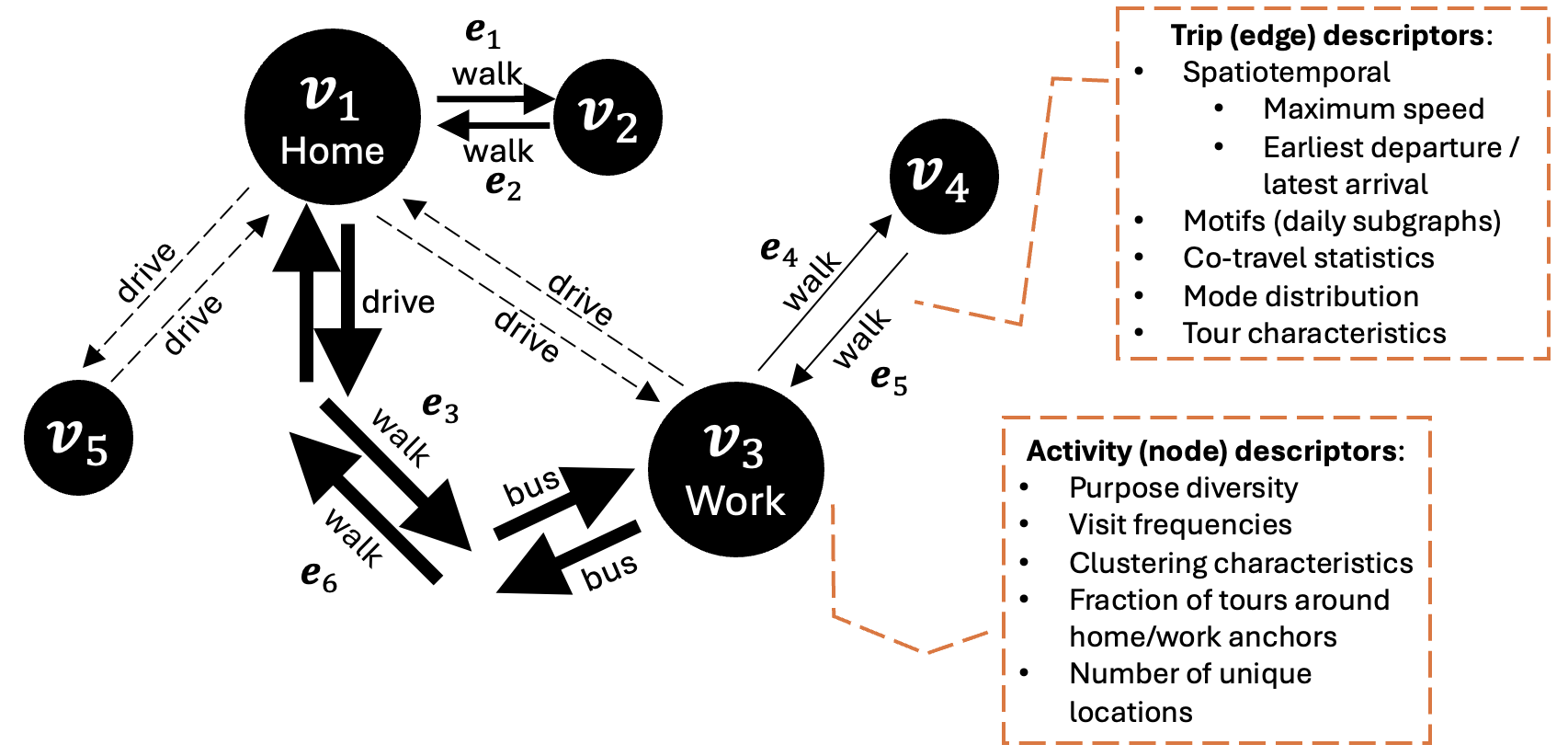}
    \caption{Example daily mobility graph where the edges are chronologically numbered. Width of trip arrows corresponds to frequency of visits over observation period. Dashed arrows denote known trips that are not observed on this day.}
    \label{fig:1}
\end{figure}

\subsubsection{Activity (node) features}
\label{sec4.1.1}
Let $\mathbf{x}=(x_1, \ldots ,x_N)$ be the frequency vector of an individual’s $N$ distinct origin-destination (OD) purpose pairs, where each pair connects two activity purposes (e.g., going home, to the gym, to school). The total trip count is $T=\sum_i x_i$. To quantify how evenly trips are distributed across these OD purpose pairs, we compute the Shannon entropy \citep{Shannon1948mathematical}:
\begin{equation}
    H_t = -\sum_{i=1}^N \frac{x_i}{T} \log_2\left(\frac{x_i}{T}\right),
\end{equation}
\noindent which is always non-negative and upper-bounded by $\log_2(N)$. High entropy indicates a diverse and balanced use of OD purpose pairs, while low entropy reflects a concentration of trips on just a few repeated purposes. While Shannon entropy measures the diversity of trip distribution, the Gini coefficient captures the degree of inequality in how frequently OD purpose pairs are used. Let the edge counts be sorted in non-decreasing order, $x_{(1)} \leq \ldots \leq x_{(N)}$, and define the cumulative trip count to rank $k$ as $C_k = \sum_{i\leq k} x_{(i)}$. Then the Gini coefficient \citep{Dorfman1979Formula} is:
\begin{equation}
    G = 1 + \frac{1}{N} - \frac{2}{NT}\sum_{k=1}^N C_k.
\end{equation}
\noindent A higher Gini indicates that a small number of purposes account for most trips, while a lower Gini suggests more equitable use across destinations. This reflects the extent to which an individual's travel is exploratory versus habitual, connecting to classical notions of travel regularity \citep{Alessandretti2018Evidence, Kitamura1987Regularity}.

To quantify global cohesion in the mobility graph, we compute the \textit{global clustering coefficient} $C_{\text{glob}}$ \citep{Opsahl2009Clustering}. This metric captures the tendency for travel purposes to be mutually connected through sequences of trips, forming tightly knit triangular structures. Let $\tau_{\triangle}$ denote the number of closed triplets (triangles) and $\tau_{\wedge}$ the number of connected triplets (wedges). Then:
\begin{equation}
    C_{\text{glob}} = \frac{3 \tau_{\triangle}}{\tau_{\wedge}}.
\end{equation}

In our context, where nodes are activity purposes and edges are trips, high clustering implies that individuals frequently travel between triplets of destinations in looping patterns (e.g., home to gym to store to home), rather than taking out-and-back trips. This metric overlaps with the concept of trip chaining, where multiple activities are linked into a single tour rather than occurring as separate out-and-back trips from a primary anchor (e.g., home or work) \citep{Ellegard1977Activity, Hanson1980importance}. To complement the global clustering coefficient, which reflects the overall cohesion of the network, we also compute the \textit{mean local clustering coefficient} $\bar{c}$ \citep{Kaiser2008Mean}. This measure averages each node's local transitivity, placing greater emphasis on peripheral or low-degree destinations:
\begin{equation}
    \bar{c} = \frac{1}{N}\sum_{v=1}^N \frac{2t_v}{k_v(k_v-1)}
\end{equation}
\noindent where $k_v$ is the degree of node $v\in V(G)$ (i.e., the number of other destinations directly connected to it), and $t_v$ is the number of triangles that include $v$. In mobility terms, a high $\bar{c}$ indicates that even less frequently visited destinations are embedded in tightly connected clusters. For example, this may suggest that auxiliary stops (e.g., a coffee shop, child’s school) are routinely integrated into a larger, cohesive tour structure rather than occurring in isolation.

\subsubsection{Trip (edge) features}
\label{sec4.1.2}

To characterize the edge layer of each mobility graph we compute three non-overlapping subfamilies of descriptors: spatiotemporal characteristics, daily travel motifs, and social co-travel mix. Let $n_{\text{trips}}$ and $n_{\text{tour}}$ denote, respectively, the number of observed trips and the number of closed tours (roundtrips anchored at either home or work) accumulated over the study horizon for a given individual. On the spatiotemporal side, we compute the fraction of trips taken during peak hours ($f_{\text{rush}}$; defined as 7–9 AM or 4–6 PM) and the fraction taken on weekends ($f_{\text{weekend}}$). We also extract the earliest, average, and latest departure times across all observed days. In addition, we calculate the average and maximum values for trip duration, speed, and distance over the study period.

The heterogeneity of daily mobility behavior tends to boil down to a handful of unique patterns, or “motifs” \citep{Schneider2013Unravelling}. Inspired by the success attained by \cite{Wu2019Inferring}, we derive motif counts after collapsing consecutive duplicate travel purposes (e.g., \textit{home} $\rightarrow$ \textit{home} $\rightarrow$ \textit{store} becomes \textit{home} $\rightarrow$ store). The resulting sequence is parsed into one or more motifs drawn from the canonical set \{single-no-return, out-and-back, chain, single-cycle, double-cycle, cycle-chain\} (see Figure 2). Let $m_j$ be the count of motifs of type $j$ accumulated over all observed days and $M=\sum_j m_j$  the individual’s total motif count. We retain the motif fractions $f_k=m_j/M$ together with the motif entropy $H_m = -\sum_j \frac{m_j}{M} \log_2(\frac{m_j}{M})$ which summarizes how evenly the six patterns are observed.

\begin{figure}[!]
    \centering
    \includegraphics[width=\linewidth]{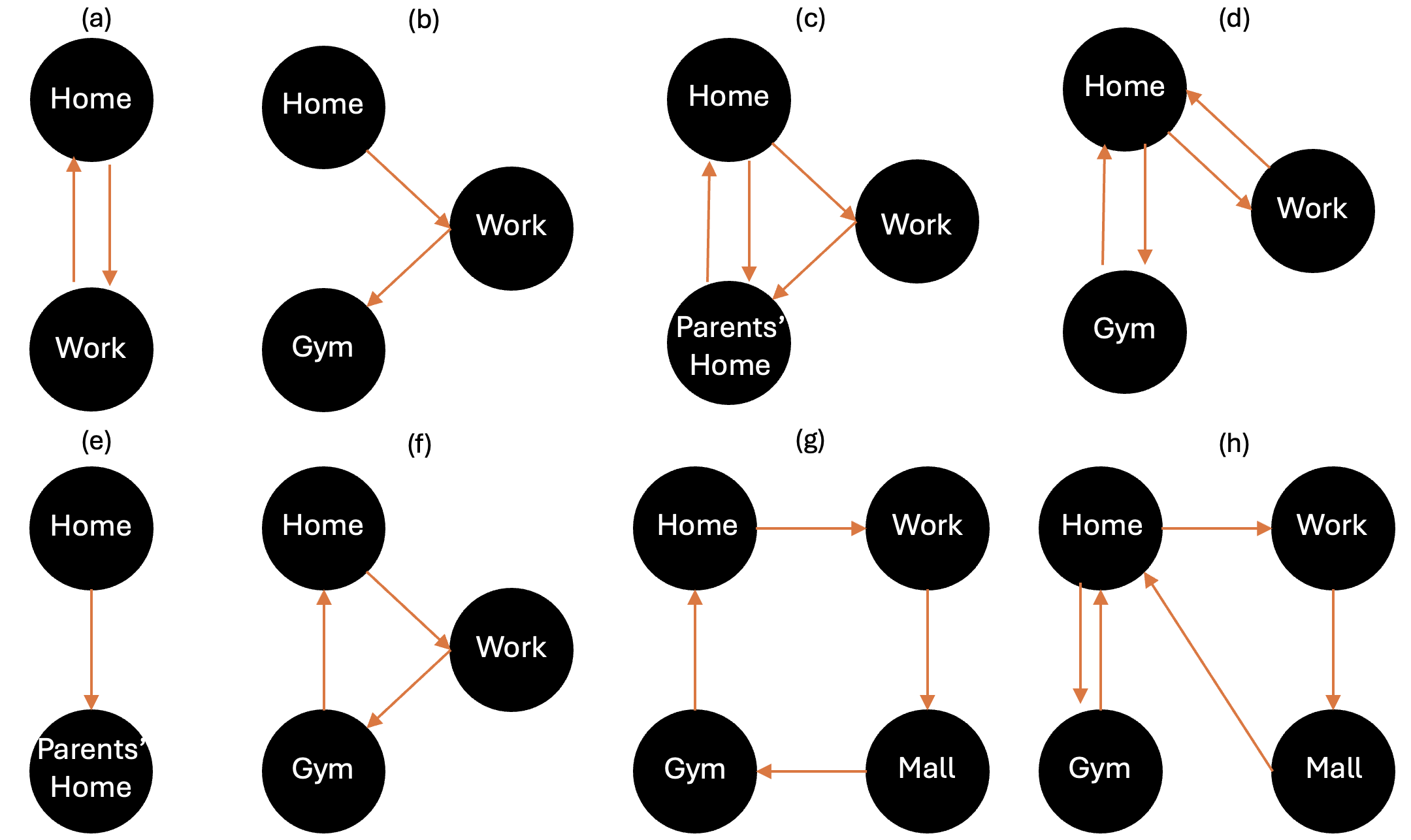}
    \caption{Illustration of daily travel motifs \citep{Schneider2013Unravelling, Wu2019Inferring}; (a) Out-and-back; (b) Chain; (c) Cycle-chain; (d, h) Double-cycle; (e) Single-no-return; (f, g) Single-Cycle}
    \label{fig:2}
\end{figure}

Trips can also be composed of multiple modes, the use of which can correlate with income and age. Let $\mathbb{1}[mm]$ be the indicator that a given tour involves at least two distinct transport modes. The \textit{multi-modal fraction} is then
\begin{equation}
    f_{mm} = \frac{\mathbb{1}[mm]}{n_{\text{tour}}}
\end{equation}
\noindent A higher value reflects broader access to, or preference for, heterogeneous mode combinations (e.g., walk-bus-walk), which we find is positively correlated with income in the Seattle context (see Figure 5).

Finally, whether one travels with companions correlates strongly with multiple sociodemographic labels. Thus, we tag each trip as \textit{solo}, \textit{with household companion(s)}, or \textit{with non-household companion(s)}. Let $n_{\text{solo}}$,  $n_{\text{hh}}$, $n_{\text{nonhh}}$ be the respective counts. We define
\begin{equation}
    f_{\text{solo}} = \frac{n_{\text{solo}}}{n_{\text{trips}}}, \quad
    f_{\text{hh}} = \frac{n_{\text{hh}}}{n_{\text{trips}}}, \quad
    f_{\text{nonhh}} = \frac{n_{\text{nonhh}}}{n_{\text{trips}}},
    \label{eq:companion-fractions}
\end{equation}
\noindent and composite fraction with companions $f_{\text{comp}}=1-f_{\text{solo}}$. These metrics embed information on household structure, caregiving roles, and wider social engagement. An illustrative computation for one travel day appears in Figure 3.

\begin{figure}[!]
    \centering
    \includegraphics[width=\linewidth]{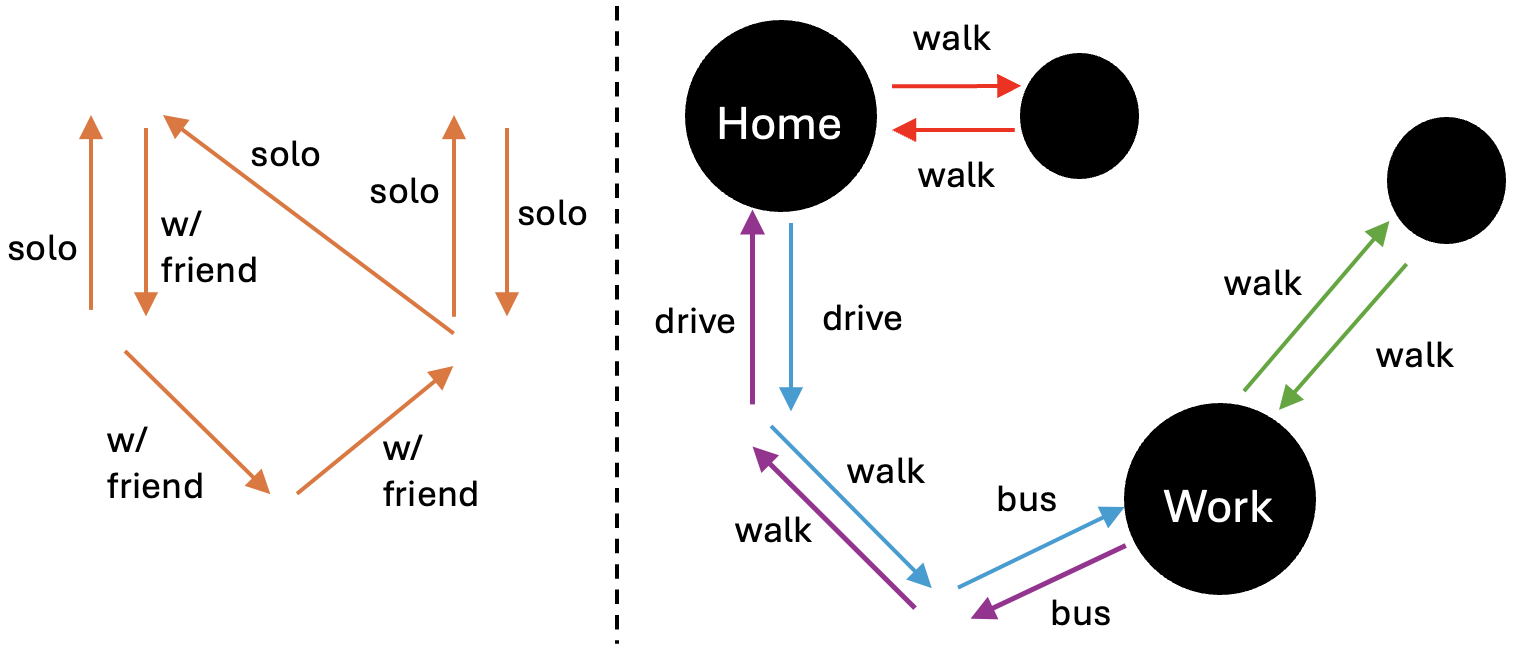}
    \caption{Illustration of selected metrics; (left) example travel day. In this case, $n_{\text{trips}}=7$ and $f_{\text{comp}}=3/7$; (right) each color denotes a tour to/from the anchors. In this case, $n_{\text{tour}}=4$ and $f_{\text{mm}}=2/4$. }
    \label{fig:3}
\end{figure}

\subsection{Probabilistic scoring and calibration for categorical targets}
\label{sec4.2}
We next tackle the retention of imputation uncertainty in supervised multi-class classification. Here, we try to answer two methodological questions: (1) How can a model provide both class probabilities and a principled measure of confidence? (2) Which evaluation criteria reward not only accuracy but also well-calibrated uncertainty? We begin with definitions and then describe metrics and visual tools to understand the output of various models. 

\subsubsection{Definitions}
\label{sec4.2.1}
Let $Y \in \{1, \dots, K\}$ denote the sociodemographic label and $X$ the mobility graph descriptors, both drawn from the ground truth joint distribution $\pi(X, Y) = \pi(Y \mid X)\pi(X)$. A model $m$ outputs $m(X) = (\hat{Y}, \hat{P})$, where $\hat{Y}$ is the predicted class and $\hat{P}$ the associated confidence (e.g., probability of correctness). We would like calibrated confidence estimates, meaning that if $m$ outputs a confidence of 0.6 for 10 predictions, roughly 6 should be correct. Formally, \textit{perfect calibration} is defined as
\begin{equation}
    \Pr(\hat{Y} = Y \mid \hat{P} = p) = p \quad \text{for all } p \in [0, 1],
    \label{eq:perfect-calibration}
\end{equation}
\noindent where the probability is over the joint distribution. A lack of calibration can lead to biased share estimates and misleading uncertainty, whereas well-calibrated models ensure that confidence values are useful and interpretable. However, the probability above cannot be computed directly from finite samples, as $\hat{P}$ is a continuous random variable. We therefore rely on empirical approximations to estimate calibration.

\subsubsection{Metrics and reliability diagrams}
\label{sec4.2.2}
We evaluate predictions with complementary measures of separability, probability quality, and calibration. Top 1 accuracy is the fraction of test cases for which the class with highest predicted probability coincides with the true label, which is interpretable as “percentage correct”. However, it evaluates predictions at a single decision rule (i.e., choose the argmax), and is therefore sensitive to class imbalance and agnostic to the quality of the full probability vector. To assess separability independent of any fixed threshold, we also report the area under the ROC curve (AUROC), which measures how often the model ranks the true class above competing classes ($0.5 = $ random, $1 =$ perfect). For multiclass tasks, we compute the one-against-rest macro-AUROC to ensure equal contribution from each class \citep{Hand2001Simple}. AUROC generalizes accuracy by integrating over all possible confidence thresholds, but it still does not assess probability calibration. A model can have high AUROC while producing poorly calibrated probabilities. Thus, we report AUROC alongside metrics that reward well-calibrated probability estimates to evaluate uncertainty quality.

One such metric is the negative log-likelihood (NLL), a standard measure of a probabilistic model’s quality \citep{Hastie2001Elements}. It is also commonly called the cross-entropy loss in the context of deep learning \citep{LeCun2015Deep}. Given $\{(x_i, y_i)\}_{i=1}^N$ as inputs and $K$ classes, a probabilistic classifier specifies a categorical distribution $q$, where each sample has predicted class probabilities $\hat{\mathbf{p}}_i = (\hat{p}_{i1}, \dots, \hat{p}_{iK})$ satisfying $\sum_k \hat{p}_{ik} = 1$. The likelihood of the observed label $y_i$ under the model for input $x_i$ is the scalar $q(y_i \mid x_i) = \hat{p}_{i, y_i}$. Then, the NLL averages the negative log of those probabilities:

\begin{equation}
\text{NLL} = -\frac{1}{N} \sum_{i=1}^N \log q(y_i \mid x_i) = -\frac{1}{N} \sum_{i=1}^N \log \hat{p}_{i, y_i}.
\label{eq:nll}
\end{equation}
\noindent NLL is minimized when the predicted probabilities match the true conditional distribution \citep{Gneiting2007Strictly}. Intuitively, NLL rewards placing high probability on the correct class and heavily penalizes overconfident mistakes.

On the other hand, the Expected Calibration Error (ECE) summarizes how well confidences match accuracies. For each sample $i$, let $\hat{y}_i = \arg\max_k \hat{p}_{ik}$ be the predicted label, and let the confidence of the prediction be the highest predicted probability: $\hat{p}_i = \max_k \hat{p}_{ik}$. To derive ECE, we partition the predictions into $M$ confidence bins (each of size $1/M$). Let $B_m$ be the set of indices of samples whose prediction confidence falls into the interval $I_m = \left(\frac{(m-1)}{M}, \frac{m}{M}\right]$. The accuracy and average confidence within bin $B_m$ are:

\begin{equation}
\text{acc}(B_m) = \frac{1}{|B_m|} \sum_{i \in B_m} \mathbb{1}(\hat{y}_i = y_i),
\label{eq:acc-bin}
\end{equation}

\begin{equation}
\text{conf}(B_m) = \frac{1}{|B_m|} \sum_{i \in B_m} \hat{p}_i,
\label{eq:conf-bin}
\end{equation}
\noindent where $y_i$ is the true class label for sample $i$. Given these definitions, the expected calibration error is the weighted average of the absolute accuracy-confidence gaps \citep{Naeini2015Obtaining}:

\begin{equation}
\text{ECE} = \sum_{m=1}^M \frac{|B_m|}{N} \left| \text{acc}(B_m) - \text{conf}(B_m) \right|.
\label{eq:ece}
\end{equation}
\noindent The difference between accuracy and confidence for a given bin is called the \emph{calibration gap}. Thus, a perfectly calibrated model attains $\text{ECE} = 0$. In our experiments, we pre-specify $M=15$ and use equal-width bins, following common practice in the calibration literature \citep{Naeini2015Obtaining}. This number provides a practical balance between resolution and statistical stability: too few bins obscure fine-grained miscalibration, while too many yield noisy estimates due to small sample counts per bin. Equal-width binning ensures that confidence intervals are evenly spaced and facilitates comparability across models and tasks. 

One related diagnostic tool we leverage is the \textit{reliability diagram} (e.g., shown in Figure~\ref{fig:7}), which visually represents model calibration. These diagrams plot expected sample accuracy (Eq.~\ref{eq:acc-bin}) as a function of confidence (Eq.~\ref{eq:conf-bin}). If the model is perfectly calibrated, the diagram should trace the identity line. Points below the diagonal indicate over-confidence, while those above indicate under-confidence. Note that reliability diagrams do not display the proportion of samples in each bin, and thus cannot be used to estimate how many samples are calibrated.

\subsection{Multitask Learning}
\label{sec4.3}
Multitask learning improves generalization by \textit{inductive transfer}, the idea that learning several related tasks together leads to better performance than learning each one in isolation \citep{Caruana1997Multitask}. In the classical ``hard parameter sharing'' formulation, tasks share part of the model’s parameters, which acts as a data-dependent regularizer that limits the effective hypothesis space. When tasks are related, this sharing reduces sample complexity: fewer labeled examples per task are needed to reach a given level of accuracy \citep{Baxter2000Model}.

Let $T$ tasks be indexed by $t \in \{1, \dots, T\}$, where each task provides samples $(X, y_t) \sim \mathcal{D}_t$ and a task-specific loss $\ell_t$. We learn a shared representation $h = g(X; \theta)$ and task-specific predictors $f_t(h; \phi_t)$ by minimizing the weighted multi-task risk:

\begin{equation}
\min_{\theta, \{\phi_t\}} \sum_{t=1}^T w_t \, \mathbb{E}_{(X, y_t) \sim \mathcal{D}_t} \left[ \ell_t\left(f_t(g(X; \theta); \phi_t), y_t \right) \right] + \Omega(\theta, \phi_t),
\label{eq:mtl-objective}
\end{equation}
\noindent where $w_t$ are task weights, $\theta$ are the shared parameters of the representation $h$, $\phi_t$ are the task-specific parameters of head $f_t$, and $\Omega$ is a standard parameter regularization term. In practice, we optimize the empirical version using mini-batches sampled from the joint dataset and backpropagate the sum of per-task losses; the shared parameters $\theta$ receive gradients from all tasks and thereby encode common structure.

\begin{figure}[!]
    \centering
    \includegraphics[width=\linewidth]{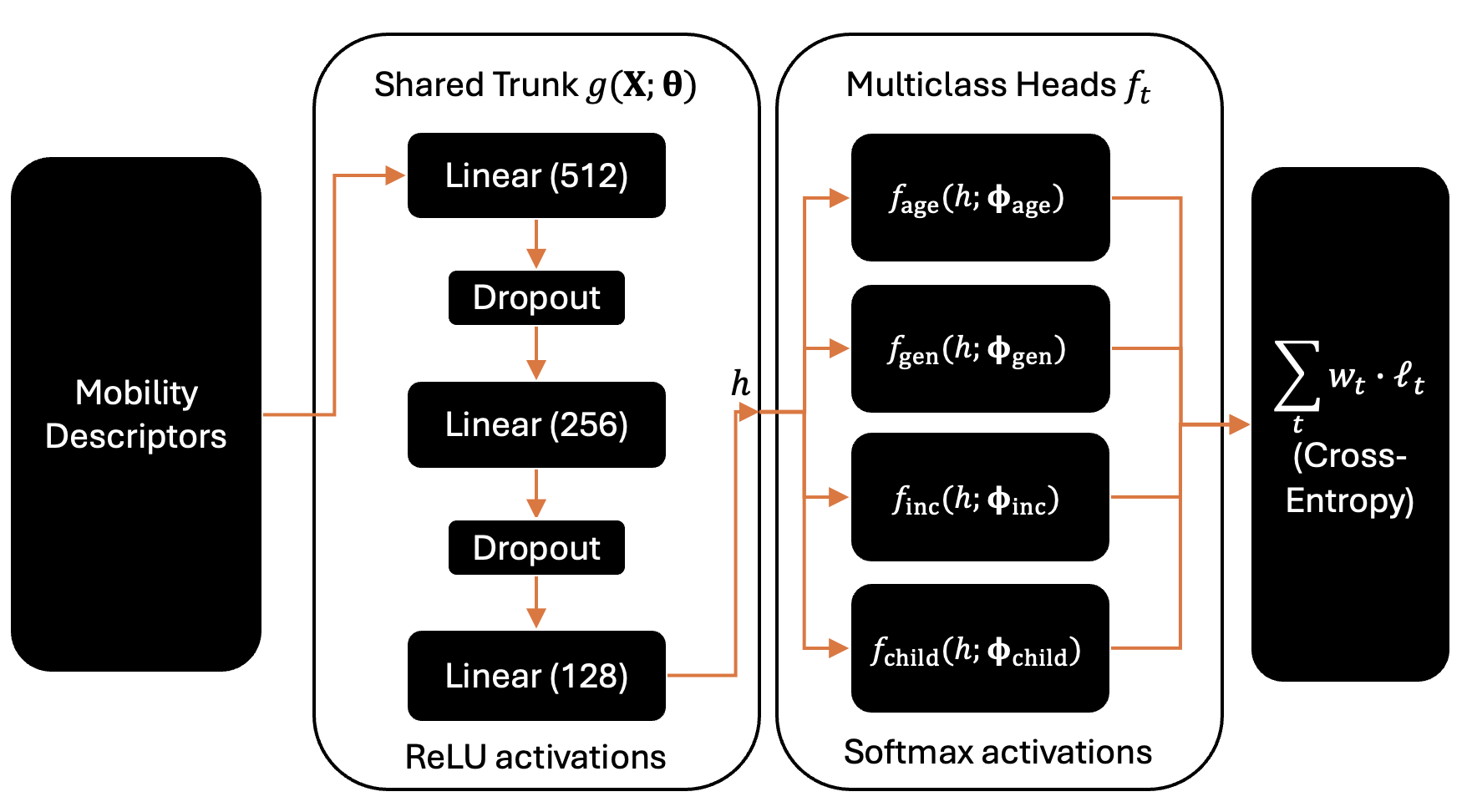}
    \caption{Shared-trunk multitask architecture. Mobility descriptors are mapped to a shared representation $h = g(X; \theta)$ by a three-layer feed-forward network with ReLU activations and dropout. Four task-specific heads $f_t$ produce class probabilities via softmax. Training minimizes the cross-entropy loss $\sum_t w_t \cdot \ell_t$ where we set the weights to be equal.}
    \label{fig:4}
\end{figure}

In this study, we adopt hard parameter sharing in a feed-forward network (shown in Figure~\ref{fig:4}). A shared trunk $g$ maps mobility features to a latent representation via three connected layers with rectified linear units (ReLU) and dropout. ReLU enables the network to learn complex nonlinear functions by zeroing out negative inputs, while dropout improves generalization by randomly disabling activations during training and encouraging distributed representations \citep{LeCun2015Deep}. Four multiclass task heads $f_t$ branch from this representation to predict age, gender, income, and number of children. This constitutes a genuine MT setup because (i) all tasks update the shared trunk during training, and (ii) only a small, task-specific set of parameters is unique to each head. The training objective is the weighted sum of task losses, where we use equal weights for each task.

The mobility descriptors given in Section~\ref{sec4.1} may capture latent routines that are jointly informative for several sociodemographic attributes. Thus, sharing this representation across the learning tasks pools statistical evidence, improving sample efficiency for sparsely labeled or imbalanced tasks, and regularizing confidence so that probabilities remain conservative under modest distributional shifts. Consistent with the theory above, our experiments show that the shared-trunk model often matches or exceeds single-task baselines in AUROC while delivering lower NLL and ECE, particularly when training data are limited or the test set differs from the training set.

\section{Experiments}
\label{sec5}
We showcase our results. In \ref{sec5.1}, we analyze correlations between our descriptors and sociodemographic targets, which fit well-known tropes in the literature. \ref{sec5.2} assesses the extent to which our features improve the predictability of demographics and model calibration. In \ref{sec5.3}, we demonstrate the value of multitask learning, which improves outcomes in data scarce regimes or on test sets that differ from the training set. 

\subsection{Linkages between mobility descriptors and sociodemographics}
\label{sec5.1}
\begin{figure}[ht]
    \centering
    \begin{subfigure}[t]{0.48\linewidth}
        \centering
        \includegraphics[width=\linewidth]{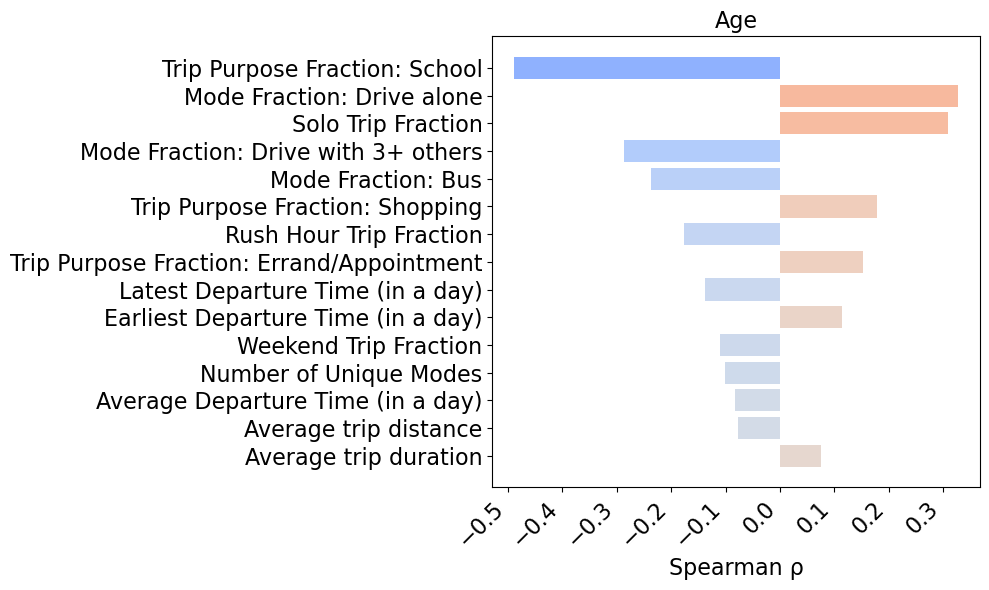}
    \end{subfigure}
    \hfill
    \begin{subfigure}[t]{0.48\linewidth}
        \centering
        \includegraphics[width=\linewidth]{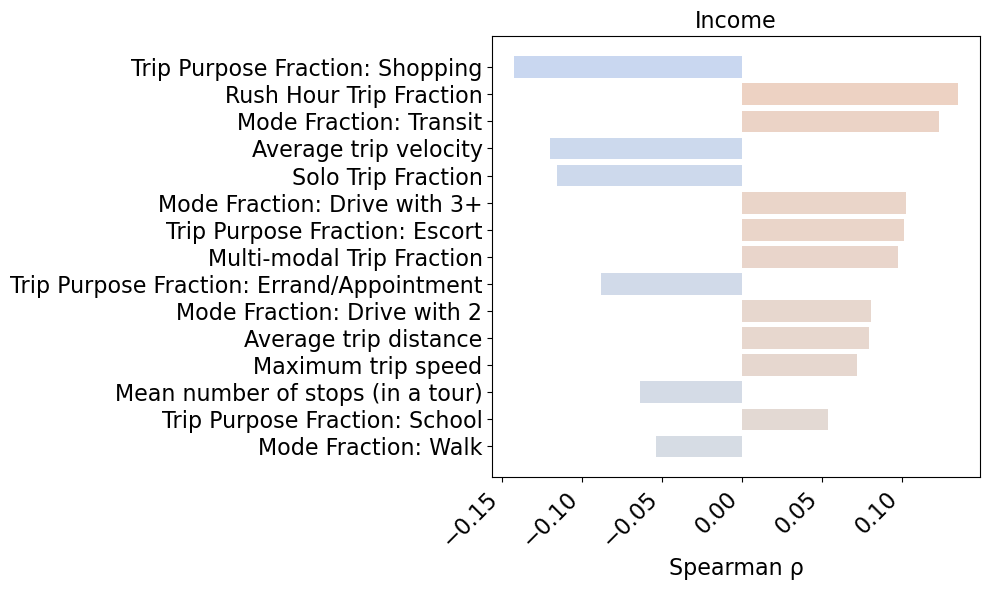}
    \end{subfigure}

    \vspace{0.3cm}

    \begin{subfigure}[t]{0.48\linewidth}
        \centering
        \includegraphics[width=\linewidth]{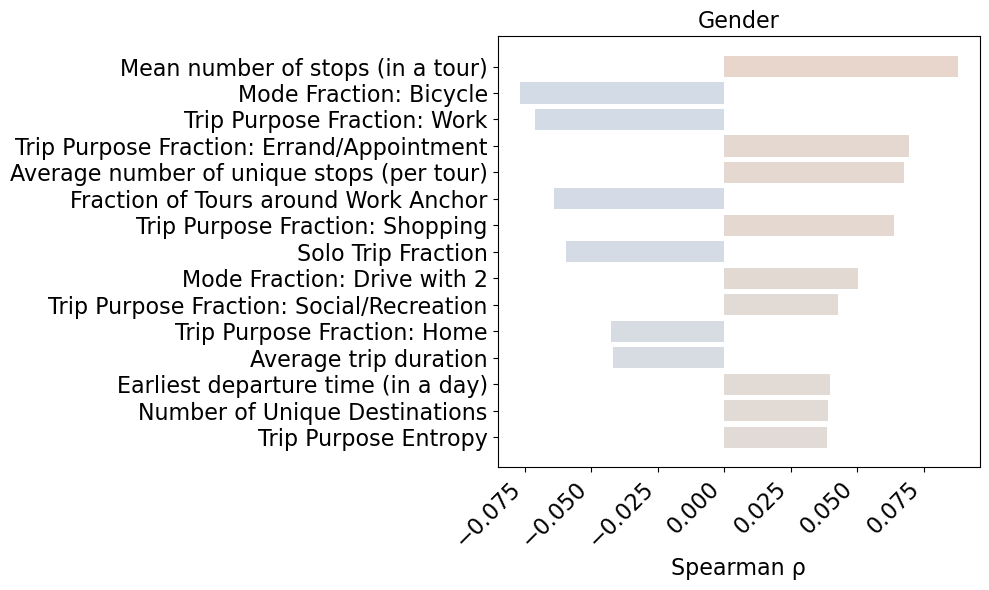}
    \end{subfigure}
    \hfill
    \begin{subfigure}[t]{0.48\linewidth}
        \centering
        \includegraphics[width=\linewidth]{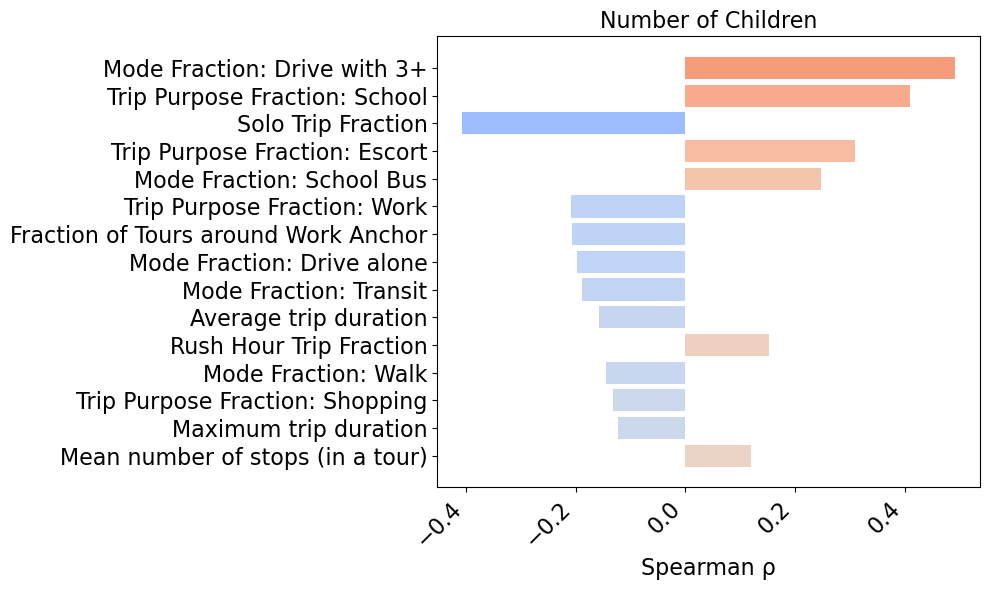}
    \end{subfigure}

    \caption{Largest magnitude Spearman rank correlations (all $p < 0.001$) between mobility descriptors and demographics. (top left): age; (top right): household income; (bottom left): gender; (bottom right): number of children. Bars to the left indicate negative associations; to the right, positive.}
    \label{fig:5}
\end{figure}

Figure \ref{fig:5} shows the largest magnitude correlations between selected sociodemographics and our feature set. Age is most strongly related to the share of school‑purpose travel ($\rho \approx -0.50$) and the drive‑alone fraction ($\rho \approx +0.30$), with younger travelers exhibiting more bus use and group car travel. Income is negatively associated with the share of shopping trips ($\rho < 0$) but positively related to trip speed, distance, and car use, consistent with longer, faster trips among higher-income travelers. Interestingly, higher income also correlates with greater transit use, likely reflecting the prevalence of white-collar commuters in Seattle who rely on transit for downtown access. Gender effects are modest ($|\rho| \leq 0.08$) but systematic: men tend to bike more, travel solo, and have a higher fraction of their trips be work-related, while women have more complex tours, more errand/appointment trips, and more shopping trips. By contrast, the number of children shows much stronger signal (up to $|\rho| \approx 0.5$): households with more children are characterized by more carpooling ($3+$ in vehicle) and higher shares of school-bus, school, and escort travel, alongside lower prevalence of solo driving, transit, and walking, fewer work-anchored tours, shorter trip durations, and fewer shopping trips. These patterns reflect the child-serving, time-constrained nature of family travel.

Table \ref{tab:linreg} shows the results of linear regression models with the sociodemographic attributes as the independent variables and our features as the dependent variables. Due to high negative correlation between age and household size (older respondents tend to live in smaller households as children move out), we try two variants of each model. Multimodality rises with age, income, and number of children, but the explained variance is small. The fraction with companions shows the clearest sociodemographic signal: it decreases with age and increases with income, being female, and number of children. By contrast, average local clustering is only weakly related to demographics, tending to be lower at higher incomes and slightly higher among women, with small effects. Out-and-back motif fractions decline with age and among women. Overall, co-travel patterns carry the most sociodemographic information in these linear models, while clustering and motif shares add subtle signals.

\begin{table}[ht]
\centering
\caption{Selected linear regression models}
\label{tab:linreg}
\setlength{\tabcolsep}{4pt}
\renewcommand{\arraystretch}{1.2}
\begin{tabular}{lc|cc|cc|cc|cc}
\multicolumn{2}{c|}{\makecell{Dependent \\ Variable}} & 
\multicolumn{2}{c|}{\makecell{Multi-modal \\ Fraction}} & 
\multicolumn{2}{c|}{\makecell{Fraction with \\ Companions}} & 
\multicolumn{2}{c|}{\makecell{Avg. Local \\ Clustering Coeff.}} & 
\multicolumn{2}{c}{\makecell{Out-and-back \\ Fraction (motif)}} \\
\hline
\multicolumn{2}{c|}{\makecell{Model / \\ Ind. Variable}} & M1 & M2 & M1 & M2 & M1 & M2 & M1 & M2 \\
\hline
Intercept      & & 0.50** & 0.53** & 0.46** & 0.61** & 0.35** & 0.35** & 0.42** & 0.41** \\
Age            & & 0.02** & 0.01** & $-$0.08** & $-$0.12** & $-$0.00 & 0.00 & $-$0.02** & $-$0.02** \\
Income         & & 0.03** & 0.03** & 0.01** & 0.02** & $-$0.00* & $-$0.01* & $-$0.00 & $-$0.01 \\
Gender         & & 0.01   & 0.01   & 0.05** & 0.06** & 0.02** & 0.02** & $-$0.02** & $-$0.02** \\
Household Size && 0.03** & ---    & 0.12** & ---    & $-$0.01 & --- & $-$0.01 & --- \\
\hline
$R^2$          && 0.01   & 0.01   & 0.22   & 0.16   & 0.002  & 0.001 & 0.004 & 0.003 \\
\end{tabular}

\medskip
{\raggedright [**, and * signify a p-value less than 0.01, and 0.05, respectively.] \par}
\end{table}

\subsection{Uplift of mobility graph features on predictability}
\label{sec5.2}
We evaluated the incremental predictive value of mobility features using nested feature sets that comprise of classical variables (C), spatiotemporal attributes (ST), diversity-related metrics (D), daily motifs (M), and co-travel statistics (CT), all discussed in Section 4.1. Classical (C) covariates include purpose shares (e.g., work, shopping, leisure), mode shares (drive, transit, walk), and simple tour statistics (e.g., number of tours and the share anchored at home/work). Spatiotemporal (ST) adds departure and arrival timing (e.g., first departure, latest arrival), trip durations, shares by period (peak/off‑peak, weekend), and basic speeds and distances. Diversity (D) captures how spread-out travel is across activities, modes, and origin-destination pairs (e.g., purpose entropy, the fraction of multi‑modal tours, and triadic closure among destinations). Motifs (M) comprise the fractions of canonical day‑level patterns (out‑and‑back, chains, cycles, etc.) together with a summary of their evenness. Co‑travel (CT) records with whom trips are taken: the shares of solo, with‑household, and with‑non‑household trips, and the overall accompanied fraction. In our setup, each set builds on the previous one--for example, the +ST set includes all classical variables plus spatiotemporal attributes, while the +CT set includes all preceding groups and adds co-travel statistics.

All experiments followed the same data protocol, which had two evaluation splits: on the \textit{overall} split, we created a 70/10/20 train/validation/test partition on the combined waves of the PSRC survey. On the \textit{cross-temporal} split, we trained and validated on the 2017 and 2019 waves and tested on the 2023 wave, a period with documented shifts in mobility behavior due to the aftermath of COVID-19 (see differences in wave composition in Table \ref{tab1}). For both splits, we used five-fold multilabel cross-validation on the pooled training and validation sets, with final performance reported on the fixed test set (either a random $20\%$ holdout or the 2023 wave, respectively).

\begin{figure}[!h]
    \centering
    \begin{subfigure}{\textwidth}
        \centering
        \includegraphics[width=\linewidth]{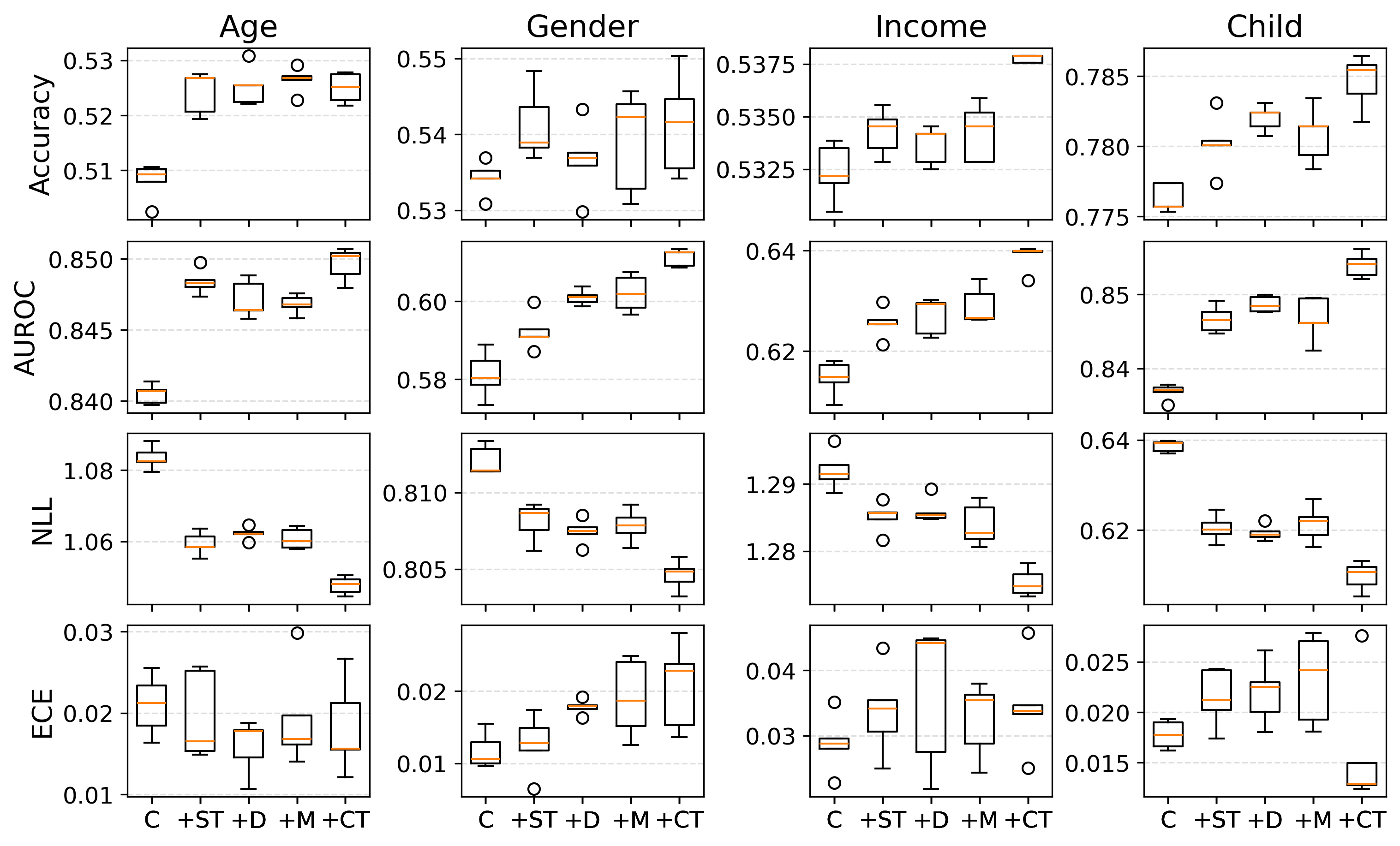}
    \end{subfigure}
    \vspace{-0.5pt}
    \begin{tikzpicture}
      \draw[dashed, line width=1pt] (0,0) -- (\linewidth,0pt);
    \end{tikzpicture}
    \vspace{-0.5pt}
    \begin{subfigure}{\textwidth}
        \centering
        \includegraphics[width=\linewidth]{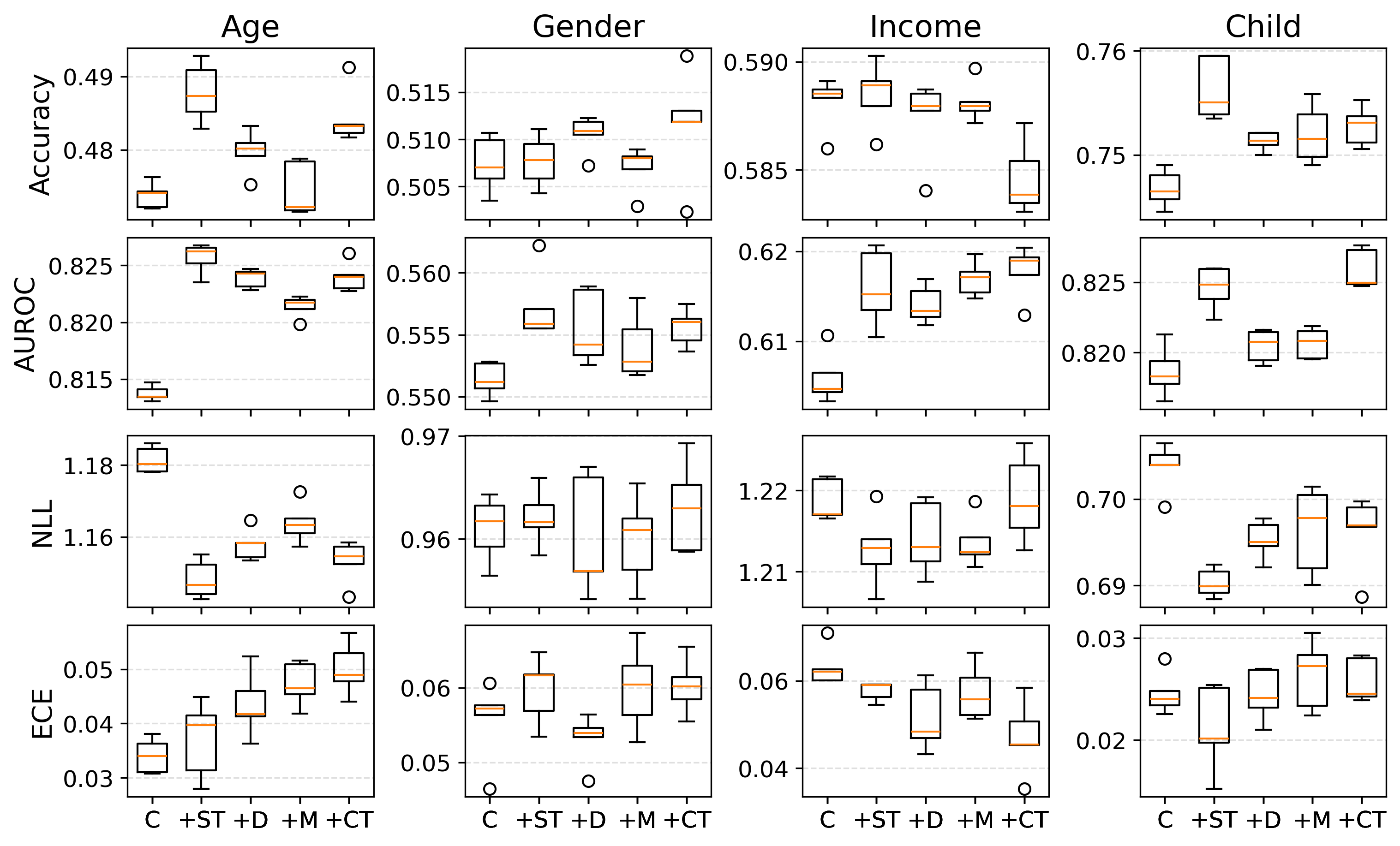}
    \end{subfigure}

    \caption{Marginal changes in top-1 accuracy (higher = better), AUROC (higher = better), NLL (lower = better), ECE (lower = better) as more features are added. (top four rows) Overall split; (bottom four rows) training with the 2017/2019 data and testing on the 2023 data.}
    \label{fig:6}
\end{figure}

We evaluated performance on top-1 accuracy, area under the receiver operating curve (AUROC), negative log-likelihood (NLL), and expected calibration error (ECE), as defined previously. Figure \ref{fig:6} presents the main results for the multi-task DNN, which is our primary model due to its compatibility with shared representation learning. To assess the robustness of the proposed feature sets across model types, we additionally evaluated three other classifiers: random forests (RF), gradient boosted machines (GBM), and support vector machines (SVM). Results for these models are reported in~\ref{app1} (Tables \ref{tab:age_overall}-\ref{tab:child_2017_2019_train_2023_test}), and show that the proposed descriptors broadly improve separability and likelihood across models. However, the most comprehensive feature set (+CT) does not always yield the best performance—approximately $44\%$ (14 out of 32) of the top metric values (shown in red) come from simpler sets. Performance trends are largely stable across model types, suggesting that the observed improvements stem primarily from the covariates rather than model-specific effects. That no single model dominates across all tasks and metrics further underscores the absence of a one-size-fits-all solution: practitioners must select models and feature sets based on their specific performance goals.

For the multi-task DNN shown in Figure \ref{fig:6}, we note the following observations of interest. First, extra covariates tend to help more under the \textit{overall} split than the \textit{cross-temporal} split. This suggests that some of the added features may capture time-specific patterns that do not generalize well when the test data come from a different distribution. In general, while richer feature sets can increase expressiveness, they may also introduce redundancy or collinearity, amplifying variance without adding meaningful new signal--especially when model capacity is not properly regularized. 

This effect is not uniform across models. As shown in Tables \ref{tab:age_overall}-\ref{tab:child_2017_2019_train_2023_test}, gradient-boosted trees (GBMs) appear less prone to overfitting under the cross-temporal split, possibly due to their implicit regularization and ability to ignore noisy or uninformative splits.

A second trend relates to model calibration. The expected calibration error (ECE) does not consistently improve with feature richness. In two of the four tasks, the most comprehensive set (+CT) yields the best-calibrated predictions, but in the others, simpler feature sets perform better. GBMs again stand out, often producing the lowest ECE scores overall, which may reflect their ability to learn conservative margins or resist overconfidence in low-signal settings. These findings highlight that while added features can increase model expressiveness, their value depends on stability across contexts and their interaction with model calibration.

Finally, the proposed descriptors generally improve both class separability (AUROC) and model fit (NLL). Under the overall split, gains tend to increase with each feature group, with the +CT set frequently yielding the best performance and rarely underperforming simpler sets. In the cross-temporal setting, improvements are more muted and task-dependent. For Age, +CT performs best for RF and GBM and is competitive for DNNs. For Number of Children, +CT achieves the highest AUROC, while +D yields the best NLL. Results for Household Income and Gender are more variable, with no single feature set dominating across metrics or models. These patterns suggest that feature utility is context- and task-dependent, with diminishing returns or instability emerging under distribution shift.

\begin{figure}[!h]
    \centering
    \begin{subfigure}[t]{0.48\linewidth}
        \centering
        \includegraphics[width=\linewidth]{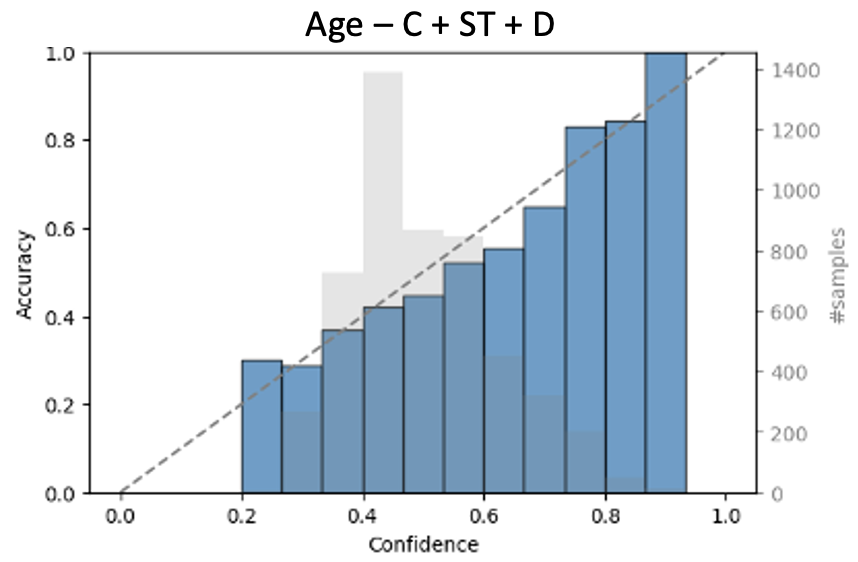}
    \end{subfigure}
    \hfill
    \begin{subfigure}[t]{0.48\linewidth}
        \centering
        \includegraphics[width=\linewidth]{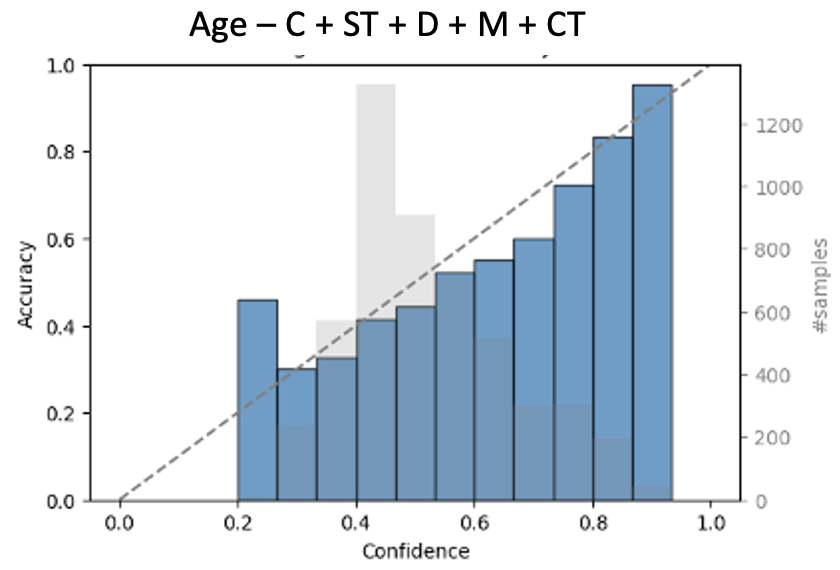}
    \end{subfigure}
    \begin{subfigure}[t]{0.48\linewidth}
        \centering
        \includegraphics[width=\linewidth]{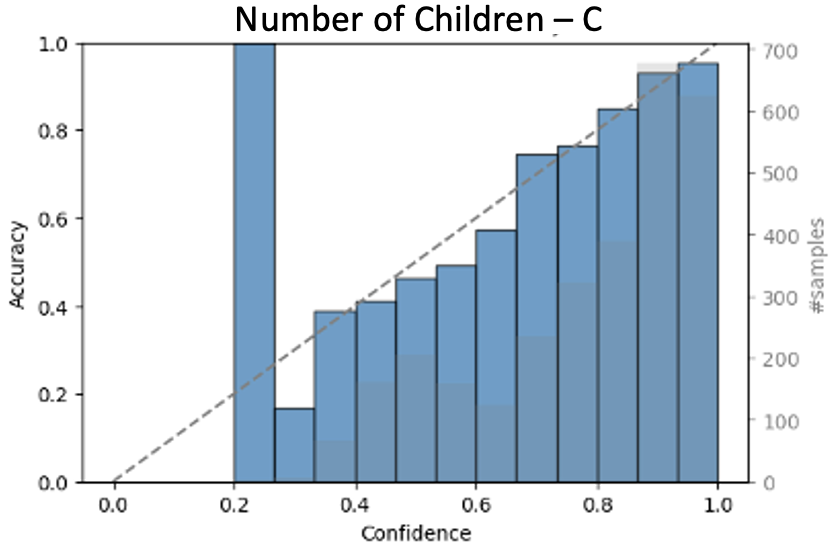}
    \end{subfigure}
    \hfill
    \begin{subfigure}[t]{0.48\linewidth}
        \centering
        \includegraphics[width=\linewidth]{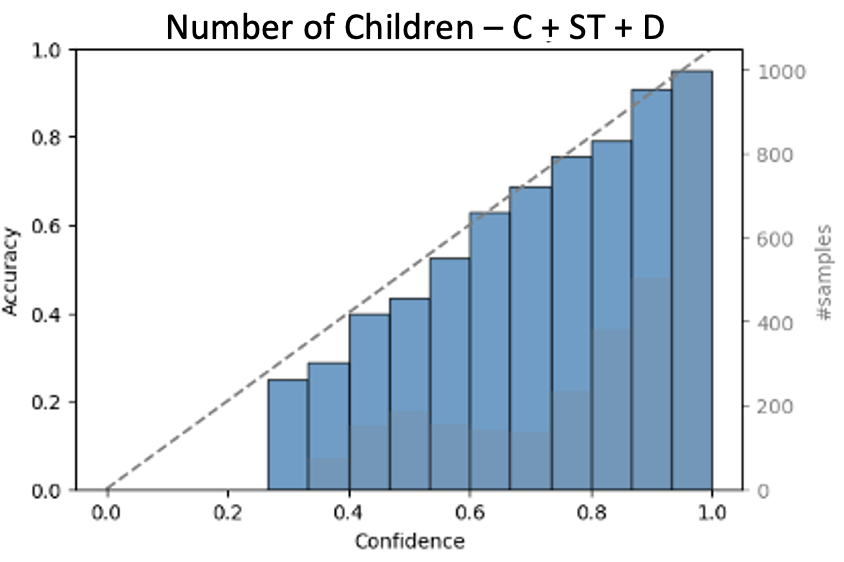}
    \end{subfigure}

    \caption{Reliability diagrams for representative settings. (TOP) Overall split, Age task with C+ST+D features (left) and All features (right). (BOTTOM) Cross-temporal split, Number of Children task with C (left) and C+ST+D (right). Bars show empirical accuracy within 15 equal-width confidence bins; the dashed line is the identity (perfect calibration). Grey histograms (right axes) give the number of samples per bin. Points above (below) the diagonal indicate under- (over-) confidence.}
    \label{fig:7}
\end{figure}

Reliability diagrams provide hints towards clarifying the variability in ECE performance. Figure \ref{fig:7} highlights that, under several settings (e.g., Age in the overall split and Number of Children under the cross-temporal split), the empirical accuracy within certain confidence bins lies well above the diagonal, indicating under-confidence: the model predicts correctly more often than its stated probabilities would suggest. This conservatism leaves AUROC and accuracy unchanged or improved but increases ECE, which penalizes the magnitude of the accuracy–confidence gap irrespective of sign. Surprisingly, this seems to happen both when more features are added (as in the top row) \emph{and} when features are removed (as in the bottom row). 

\subsection{Impact of Multitask Learning}
\label{sec5.3}

We compared a shared trunk multitask (MT) network with matched single task variants (STV) across age, gender, income, and number of children, while holding architecture, optimization, and regularization fixed (see~\ref{app1} for full hyperparameter tuning details). To stabilize learning and reduce task interference, we optionally applied a \textit{per-task layer normalization} module before each task head, normalizing activations separately for each prediction branch rather than sharing normalization statistics across tasks. To probe sample efficiency, we randomly subsampled the training split to fractions $\{1.0, 0.1 ,0.01, 0.001\}$ and evaluated on the untouched validation and test sets. To avoid overfitting as training data size got small, we used only the classical and spatiotemporal features. 

\begin{figure}[!]
    \centering
    \begin{subfigure}{\textwidth}
        \centering
        \includegraphics[width=\linewidth]{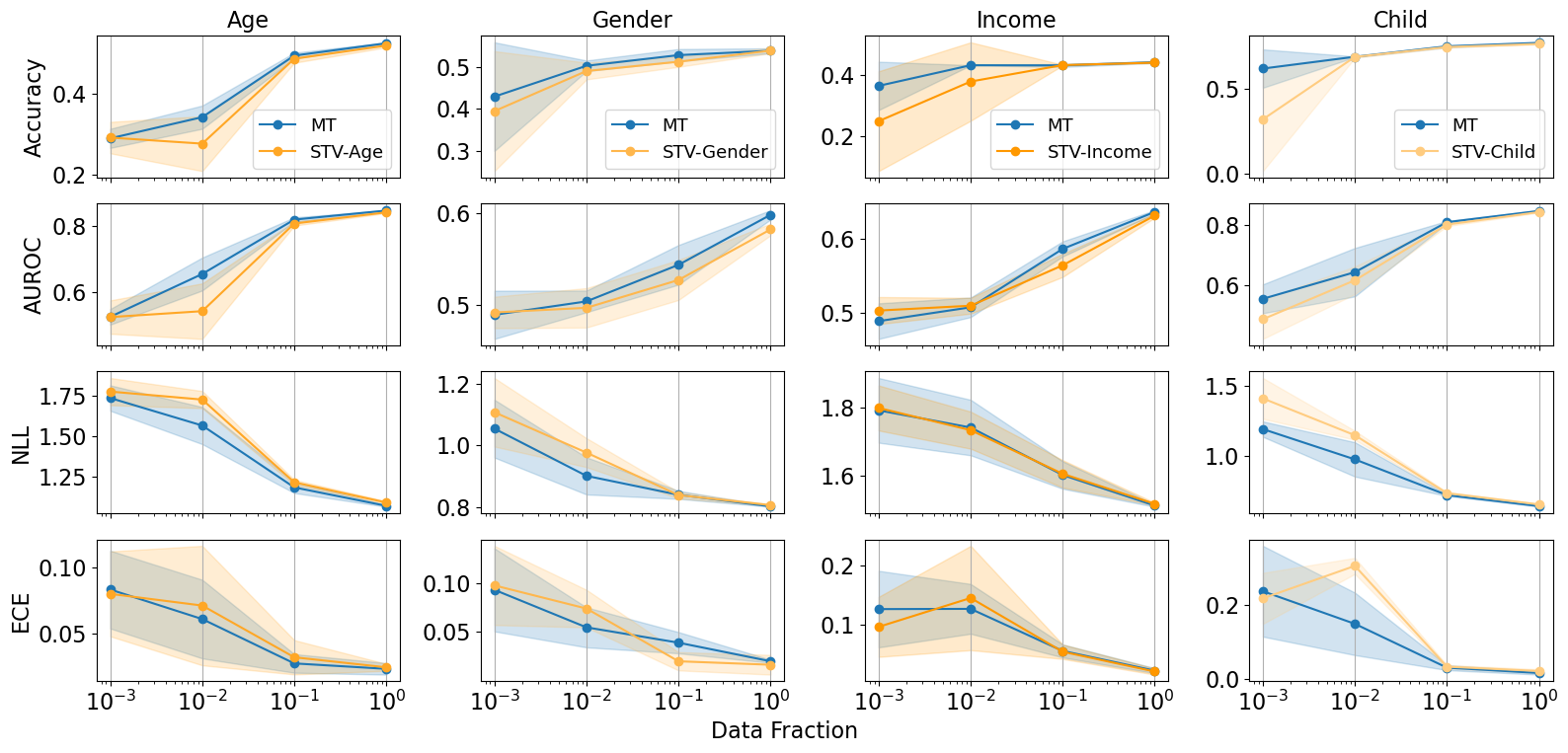}
    \end{subfigure}
    \vspace{-0.5pt}
    \begin{tikzpicture}
      \draw[dashed, line width=1pt] (0,0) -- (\linewidth,0pt);
    \end{tikzpicture}
    \vspace{-0.5pt}
    \begin{subfigure}{\textwidth}
        \centering
        \includegraphics[width=\linewidth]{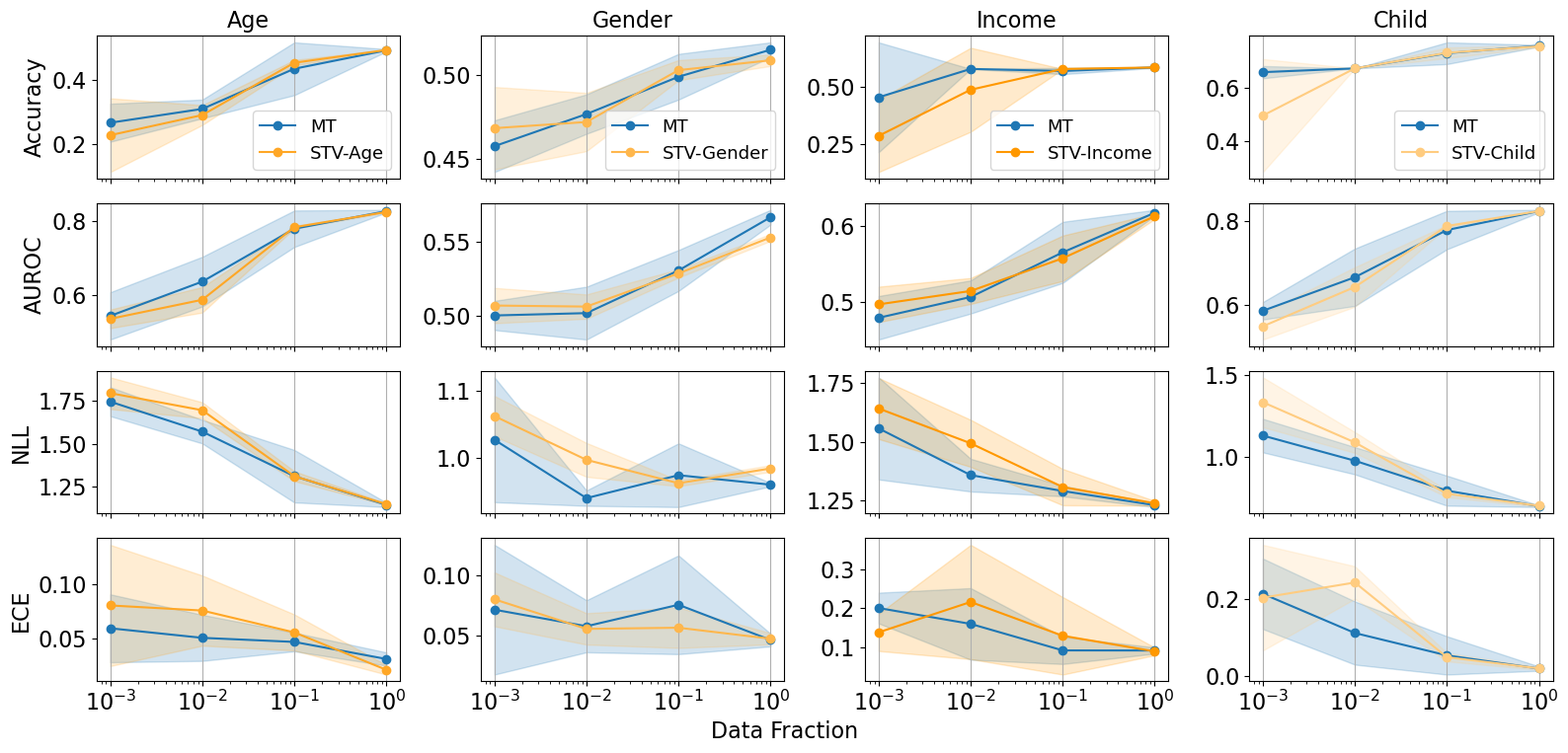}
    \end{subfigure}

    \caption{Performance of the MT variant (in blue) compared to ST variants (in orange) at different fractions of training data. (top four rows) Overall split; (bottom four rows) training with the 2017/2019 data and testing on the 2023 data.}
    \label{fig:8}
\end{figure}
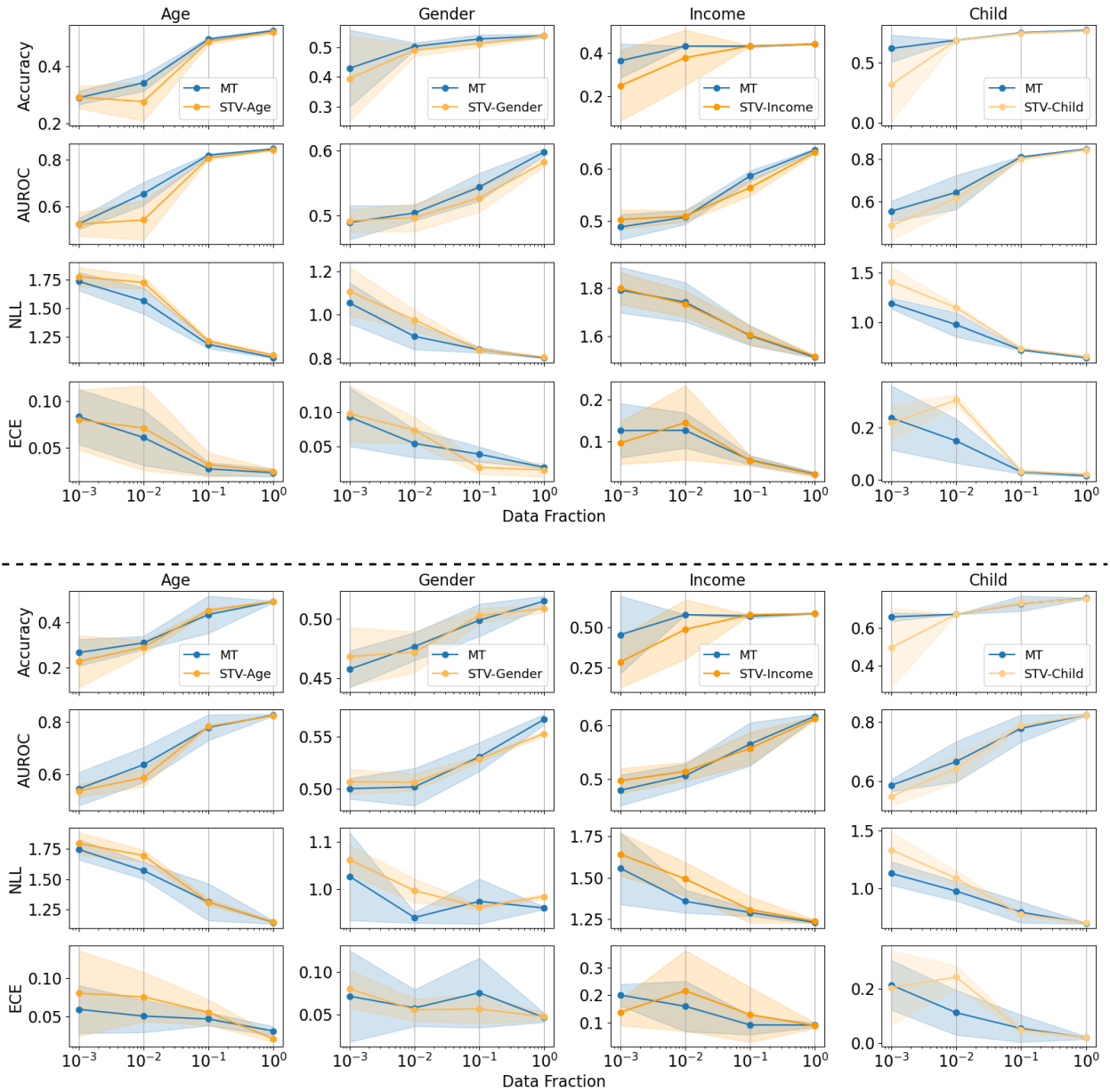

Figure \ref{fig:8} displays our findings. On the overall split (top four rows of Figure \ref{fig:8}), the MT network and and its STVs achieve comparable top-1 accuracies and AUROC at full data. As the training fraction shrinks, the relative patterns diverge by task. Age, gender, and number of children exhibit a modest MT advantage in accuracy and AUROC at intermediate and low fractions, indicating effective transfer of shared structure when labels are scarce. By contrast, HH income shows mixed patterns, particularly at the smallest fractions, suggesting that shared representation can sometimes blur task-specific distinctions (i.e., negative transfer). In these cases, the single-task networks appear better able to specialize when the shared structure among targets is weak.

The uncertainty metrics favor the multitask learning-based model. NLL remains lower for MT than STVs as data become scarce, and ECE is lower for MT for three of the four fractions over each sociodemographic target. This indicates that even when discrimination gains are inconsistent, the shared representation can still regularize learning by preventing overconfidence and smoothing probability estimates.


Under the cross-temporal generalization setting (bottom four rows of Figure \ref{fig:8}), the MT network is again competitive with STVs in accuracy and AUROC across most fractions, though the advantages are uneven. Gains are most apparent for HH income, number of children, and to a lesser extent, age, while gender shows little benefit or mild negative transfer. The improvements in NLL are more pronounced than in the overall split, suggesting that multitask learning helps regularize against overconfidence under distribution shift. By pooling representational strength across targets, the model can better withstand changes in the marginal distribution of mobility features. This effect is especially helpful when some individual targets have relatively few informative examples in the new distribution (i.e., with rare subgroups or behaviors that shift over time). This stabilizing effect is especially useful in real-world deployment, where shifts in behavior over time (e.g., due to pandemics, fuel prices, or policy changes) can erode model reliability.

\section{Conclusion}
\label{sec6}
This study advances sociodemographic inference from mobility traces along three fronts: (i) it introduces a behaviorally grounded family of higher-order mobility descriptors that move beyond first-order counts to encode trip sequencing, cohesion, and social co-travel, among other characteristics; (ii) it operationalizes uncertainty-aware evaluation for multi-class prediction in this context, a dimension largely absent in prior studies, which tend to focus on point estimates without assessing the reliability of model confidence, and; (iii) it examines how a shared-trunk multitask (MT) architecture compares to matched single-task variants in data efficiency and calibration quality. Empirically, the proposed descriptors generally raise out-of-sample accuracy and lower NLL across attributes. The benefits of MT learning, however, are nuanced. While it often improves calibration and robustness under data scarcity and temporal distribution shift, these gains are not uniform across targets. In some cases—particularly for attributes with weaker behavioral overlap or higher label noise—task interference leads to mild negative transfer, causing MT to underperform specialized single-task networks. Overall, shared representations can regularize predictions and enhance reliability, but their value depends on the degree of cross-task relatedness and the balance between shared and task-specific learning.

This work has limitations that motivate future research. We do not derive features directly from raw GPS/LBS signals; instead, we rely on processed trip diaries with purpose, mode, and co-travel labels. Many descriptors could in principle be computed from passive traces (e.g., reverse-geocoded activity locations, dwell-time anchors, and mode inference from speed/acceleration and network context), but their fidelity will depend on upstream imputation methods \citep{Gao2024Activity, Merikhipour2024Transportation}. Furthermore, our cross-temporal evaluation is confined to a single region; a natural extension is a cross-city study to assess geographic portability. 

More broadly, this paper does not take a model- or architecture-centric view of predictive performance. Our emphasis is on deriving behaviorally grounded features, quantifying their contributions, and examining model uncertainty rather than pursuing architectural novelty. The only modeling variation we explore is multitask learning, chosen to test whether shared representations help under data sparsity (i.e., building on authors' previous findings in \cite{ugurel_correcting_2024, ugurel_learning_2024}). This is a well-established line of inquiry rather than a new algorithmic proposal. That said, there is considerable potential in moving beyond hand-crafted descriptors toward models that can automatically discover structure in large-scale mobility data. Recent advances in Transformer-based and self-supervised architectures point to promising directions, particularly for unlabeled PCM with long temporal depth, where rich behavioral regularities could be captured through pretraining \citep{wu2024pretrained} and later adapted to sociodemographic inference.

\section*{Declaration of Conflicting Interests}
The authors declared no potential conflicts of interest with respect to the research, authorship, and/or publication of this article.

\section*{Funding Sources}
This research was supported by two sources: (1) the Valle Scholarship \& Scandinavian Exchange Program at the University of Washington; and (2) the Center for Understanding Future Travel Behavior and Demand (TBD), a National University Transportation Center sponsored by the US Department of Transportation under grant 69A3552344815 and 69A3552348320. 

\section*{Declaration of generative AI use}
During the preparation of this work, the authors used ChatGPT for grammar check and editing. After using this tool/service, the author(s) reviewed and edited the content as needed and take(s) full responsibility for the content of the published article.
\newpage
\bibliographystyle{elsarticle-harv} 
\bibliography{cas-refs}

\newpage
\begin{appendix}
    
\section{Experimental Details and Full Model Results}
\label{app1}

We detail the experimental setup for the rest of the classifiers used in the two experiments discussed in the main text. Both the multi-task DNNs and their single-task counterparts were built with two hidden layers, each separated by ReLU activations, and a uniform dropout rate of $0.3$ between layers. For optimization, we performed a grid search over learning rates \(\{10^{-3}, 10^{-4}, 5\times10^{-5}, 10^{-5}\}\), batch sizes \(\{16, 32, 64, 128\}\), and weight decay values \(\{10^{-3}, 10^{-4}, 10^{-5}\}\). The best configuration, determined via validation loss, used a learning rate of \(5\times10^{-5}\), batch size of \(64\), and weight decay of \(10^{-4}\). Models were trained for up to \(200\) epochs with early stopping if the validation loss did not improve for \(20\) consecutive epochs.

To isolate the effect of parameter sharing (MT) from raw capacity, we matched aggregate hidden width across conditions. The unified MT model used a shared trunk of $256 \xrightarrow[]{} 128$ units, while each single-task variant (STV) used $64 \xrightarrow[]{} 32$ units. Since there are four STVs, their combined width ($4 \times 64, 4 \times 32$) is comparable to the MT trunk ($256, 128$), making representational capacity roughly parity-matched while differing in whether features are learned jointly or separately.

\begin{figure}[!h]
    \centering
    \includegraphics[width=0.75\linewidth]{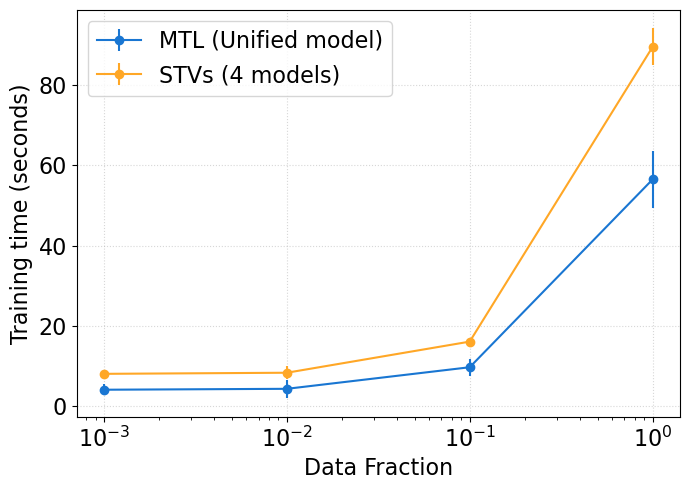}
    \caption{Comparison of training times between the unified multi-task learning (MT) model and four separate single-task variants (STVs) across varying data fractions (log-scaled on x-axis). Despite its larger architecture, the MT model trains faster overall because its shared trunk amortizes computation across tasks, whereas STVs require four independent forward–backward passes. Error bars show the standard deviation across cross-validation folds.}
    \label{fig:training_times}
\end{figure}

Despite the smaller per-model widths, STVs require training four independent networks (four forward/backward passes, four optimizers, four early-stopping loops) and thus incur higher wall-clock time than the single unified MT model, which amortizes the trunk compute across tasks. Empirically, STVs were consistently slower than MT across data fractions (see Figure \ref{fig:training_times}).

For the benchmark models leveraged in the \emph{uplift} experiments, we used \textsc{sklearn}'s built-in \hyperlink{https://scikit-learn.org/stable/modules/generated/sklearn.ensemble.RandomForestClassifier.html}{random forest (RF) classifier}, \hyperlink{https://scikit-learn.org/stable/modules/generated/sklearn.ensemble.GradientBoostingClassifier.html}{gradient boosting (GB) classifier}, and \hyperlink{https://scikit-learn.org/stable/modules/generated/sklearn.svm.SVC.html}{C-Support Vector (SVC) classifier}. For RFs and SVCs, we performed a small grid search over the number of estimators \(\{100, 300, 500\}\) and regularization parameter \(C \in \{0.1, 1, 10\}\), respectively. We selected \(500\) estimators for RFs and \(C=1\) for SVCs based on validation performance. Due to computational load, we kept the number of GB estimators at $100$. We also enabled probability estimates for the SVC classifier, which slowed down convergence. The rest of this section contains the full results tables in \ref{tab:age_overall} until \ref{tab:child_2017_2019_train_2023_test}. 

\subsection{Overall Split}
\begin{table}[H]
\caption{Performance across feature sets and models for Age (overall split). Values are mean±sd across folds; best per model in \textbf{bold}; best in metric in \textcolor{red}{\textbf{red}}.}
\label{tab:age_overall}
\begin{tabular}{llcccc}
\toprule
 &  & DNN & RF & GBM & SVM \\
Metric & Feature set &  &  &  &  \\
\midrule
\multirow{5}{*}{\rotatebox[origin=c]{90}{Accuracy}} & C & 0.508 ± 0.003 & 0.483 ± 0.003 & 0.503 ± 0.000 & 0.504 ± 0.000 \\
 & +ST & 0.524 ± 0.004 & 0.516 ± 0.004 & 0.513 ± 0.000 & 0.525 ± 0.001 \\
 & +D & 0.525 ± 0.004 & 0.517 ± 0.003 & 0.515 ± 0.000 & 0.523 ± 0.000 \\
 & +M & \textbf{0.526 ± 0.002} & 0.518 ± 0.003 & 0.514 ± 0.000 & 0.520 ± 0.000 \\
 & +CT & 0.525 ± 0.003 & \textbf{0.520 ± 0.003} & \textbf{0.518 ± 0.000} & \textcolor{red}{\textbf{0.527 ± 0.001}} \\
\cline{1-6}
\multirow{5}{*}{\rotatebox[origin=c]{90}{AUROC}} & C & 0.840 ± 0.001 & 0.815 ± 0.000 & 0.835 ± 0.000 & 0.824 ± 0.000 \\
 & +ST & 0.848 ± 0.001 & 0.834 ± 0.001 & 0.842 ± 0.000 & 0.838 ± 0.000 \\
 & +D & 0.847 ± 0.001 & 0.834 ± 0.001 & 0.842 ± 0.000 & 0.838 ± 0.000 \\
 & +M & 0.847 ± 0.001 & 0.834 ± 0.001 & 0.842 ± 0.000 & 0.836 ± 0.000 \\
 & +CT & \textcolor{red}{\textbf{0.850 ± 0.001}} & \textbf{0.837 ± 0.001} & \textbf{0.843 ± 0.000} & \textbf{0.838 ± 0.000} \\
\cline{1-6}
\multirow{5}{*}{\rotatebox[origin=c]{90}{NLL}} & C & 1.084 ± 0.003 & 1.277 ± 0.007 & 1.099 ± 0.000 & 1.134 ± 0.000 \\
 & +ST & 1.059 ± 0.003 & 1.117 ± 0.006 & 1.080 ± 0.000 & 1.096 ± 0.000 \\
 & +D & 1.062 ± 0.002 & 1.118 ± 0.003 & 1.081 ± 0.000 & 1.096 ± 0.000 \\
 & +M & 1.061 ± 0.003 & 1.120 ± 0.006 & 1.081 ± 0.000 & 1.099 ± 0.000 \\
 & +CT & \textcolor{red}{\textbf{1.048 ± 0.002}} & \textbf{1.098 ± 0.003} & \textbf{1.071 ± 0.000} & \textbf{1.086 ± 0.000} \\
\cline{1-6}
\multirow{5}{*}{\rotatebox[origin=c]{90}{ECE}} & C & 0.021 ± 0.004 & 0.060 ± 0.004 & 0.028 ± 0.001 & 0.034 ± 0.001 \\
 & +ST & 0.020 ± 0.005 & \textbf{0.029 ± 0.003} & 0.016 ± 0.001 & 0.027 ± 0.002 \\
 & +D & \textbf{0.016 ± 0.003} & 0.032 ± 0.006 & 0.017 ± 0.000 & 0.022 ± 0.003 \\
 & +M & 0.019 ± 0.006 & 0.034 ± 0.008 & 0.017 ± 0.000 & \textbf{0.021 ± 0.001} \\
 & +CT & 0.018 ± 0.006 & 0.034 ± 0.003 & \textcolor{red}{\textbf{0.016 ± 0.000}} & 0.023 ± 0.002 \\
\cline{1-6}
\bottomrule
\end{tabular}
\end{table}

\begin{table}[H]
\caption{Performance across feature sets and models for Gender (overall split). Values are mean±sd across folds; best per model in \textbf{bold}; best in metric in \textcolor{red}{\textbf{red}}.}
\label{tab:gender_overall}
\begin{tabular}{llcccc}
\toprule
 &  & DNN & RF & GBM & SVM \\
Metric & Feature set &  &  &  &  \\
\midrule
\multirow{5}{*}{\rotatebox[origin=c]{90}{Accuracy}} & C & 0.534 ± 0.002 & 0.519 ± 0.002 & 0.528 ± 0.000 & 0.530 ± 0.001 \\
 & +ST & 0.541 ± 0.005 & 0.541 ± 0.002 & 0.547 ± 0.000 & 0.538 ± 0.000 \\
 & +D & 0.537 ± 0.005 & 0.541 ± 0.006 & 0.543 ± 0.000 & 0.541 ± 0.000 \\
 & +M & 0.539 ± 0.007 & \textbf{0.541 ± 0.004} & 0.545 ± 0.000 & 0.537 ± 0.000 \\
 & +CT & \textbf{0.541 ± 0.007} & 0.540 ± 0.006 & \textcolor{red}{\textbf{0.549 ± 0.000}} & \textbf{0.546 ± 0.001} \\
\cline{1-6}
\multirow{5}{*}{\rotatebox[origin=c]{90}{AUROC}} & C & 0.581 ± 0.006 & 0.558 ± 0.002 & 0.588 ± 0.000 & 0.588 ± 0.001 \\
 & +ST & 0.592 ± 0.005 & 0.587 ± 0.004 & 0.606 ± 0.000 & 0.588 ± 0.000 \\
 & +D & 0.601 ± 0.002 & 0.591 ± 0.006 & 0.628 ± 0.000 & 0.596 ± 0.000 \\
 & +M & 0.602 ± 0.005 & \textbf{0.594 ± 0.003} & \textcolor{red}{\textbf{0.631 ± 0.000}} & 0.608 ± 0.000 \\
 & +CT & \textbf{0.611 ± 0.002} & 0.593 ± 0.007 & 0.627 ± 0.000 & \textbf{0.612 ± 0.000} \\
\cline{1-6}
\multirow{5}{*}{\rotatebox[origin=c]{90}{NLL}} & C & 0.812 ± 0.001 & 0.929 ± 0.002 & 0.813 ± 0.000 & 0.814 ± 0.000 \\
 & +ST & 0.808 ± 0.001 & 0.832 ± 0.002 & 0.806 ± 0.000 & 0.810 ± 0.000 \\
 & +D & 0.807 ± 0.001 & 0.831 ± 0.003 & 0.802 ± 0.000 & 0.809 ± 0.000 \\
 & +M & 0.808 ± 0.001 & \textbf{0.829 ± 0.000} & 0.802 ± 0.000 & 0.808 ± 0.000 \\
 & +CT & \textbf{0.805 ± 0.001} & 0.829 ± 0.002 & \textcolor{red}{\textbf{0.801 ± 0.000}} & \textbf{0.806 ± 0.000} \\
\cline{1-6}
\multirow{5}{*}{\rotatebox[origin=c]{90}{ECE}} & C & \textbf{0.012 ± 0.002} & 0.077 ± 0.002 & 0.019 ± 0.000 & 0.008 ± 0.001 \\
 & +ST & 0.013 ± 0.004 & 0.040 ± 0.006 & 0.013 ± 0.000 & \textcolor{red}{\textbf{0.005 ± 0.001}} \\
 & +D & 0.018 ± 0.001 & 0.041 ± 0.007 & 0.015 ± 0.000 & 0.010 ± 0.001 \\
 & +M & 0.019 ± 0.005 & \textbf{0.035 ± 0.003} & 0.014 ± 0.001 & 0.006 ± 0.001 \\
 & +CT & 0.021 ± 0.006 & 0.039 ± 0.004 & \textbf{0.013 ± 0.000} & 0.013 ± 0.002 \\
\cline{1-6}
\bottomrule
\end{tabular}
\end{table}

\begin{table}[H]
\caption{Performance across feature sets and models for HH Income (overall split). Values are mean±sd across folds; best per model in \textbf{bold}; best in metric in \textcolor{red}{\textbf{red}}.}
\label{tab:income_overall}
\begin{tabular}{llcccc}
\toprule
 &  & DNN & RF & GBM & SVM \\
Metric & Feature set &  &  &  &  \\
\midrule
\multirow{5}{*}{\rotatebox[origin=c]{90}{Accuracy}} & C & 0.532 ± 0.001 & 0.528 ± 0.002 & 0.535 ± 0.000 & 0.532 ± 0.000 \\
 & +ST & 0.534 ± 0.001 & 0.561 ± 0.001 & 0.531 ± 0.000 & 0.535 ± 0.001 \\
 & +D & 0.534 ± 0.001 & \textcolor{red}{\textbf{0.563 ± 0.001}} & 0.534 ± 0.000 & 0.536 ± 0.001 \\
 & +M & 0.534 ± 0.001 & 0.561 ± 0.001 & 0.532 ± 0.000 & 0.536 ± 0.001 \\
 & +CT & \textbf{0.538 ± 0.000} & 0.561 ± 0.001 & \textbf{0.536} ± 0.000 & \textbf{0.540 ± 0.001} \\
\cline{1-6}
\multirow{5}{*}{\rotatebox[origin=c]{90}{AUROC}} & C & 0.615 ± 0.003 & 0.617 ± 0.001 & 0.612 ± 0.000 & 0.599 ± 0.001 \\
 & +ST & 0.626 ± 0.003 & 0.677 ± 0.001 & 0.638 ± 0.000 & 0.626 ± 0.000 \\
 & +D & 0.627 ± 0.004 & 0.678 ± 0.002 & 0.639 ± 0.000 & 0.630 ± 0.000 \\
 & +M & 0.629 ± 0.004 & 0.678 ± 0.002 & 0.640 ± 0.000 & 0.631 ± 0.000 \\
 & +CT & \textbf{0.639 ± 0.003} & \textcolor{red}{\textbf{0.691 ± 0.001}} & \textbf{0.650 ± 0.000} & \textbf{0.644 ± 0.000} \\
\cline{1-6}
\multirow{5}{*}{\rotatebox[origin=c]{90}{NLL}} & C & 1.292 ± 0.003 & 1.556 ± 0.013 & 1.289 ± 0.000 & 1.305 ± 0.000 \\
 & +ST & 1.285 ± 0.002 & 1.216 ± 0.001 & 1.275 ± 0.000 & 1.291 ± 0.000 \\
 & +D & 1.286 ± 0.002 & 1.216 ± 0.002 & 1.273 ± 0.000 & 1.288 ± 0.000 \\
 & +M & 1.284 ± 0.003 & 1.217 ± 0.003 & 1.272 ± 0.000 & 1.287 ± 0.000 \\
 & +CT & \textbf{1.275 ± 0.002} & \textcolor{red}{\textbf{1.202 ± 0.002}} & \textbf{1.263 ± 0.000} & \textbf{1.275 ± 0.000} \\
\cline{1-6}
\multirow{5}{*}{\rotatebox[origin=c]{90}{ECE}} & C & \textbf{0.029 ± 0.004} & 0.048 ± 0.002 & 0.029 ± 0.000 & \textbf{0.025 ± 0.001} \\
 & +ST & 0.034 ± 0.007 & \textcolor{red}{\textbf{0.020 ± 0.003}} & \textbf{0.022 ± 0.000} & 0.041 ± 0.004 \\
 & +D & 0.037 ± 0.011 & 0.022 ± 0.003 & 0.031 ± 0.001 & 0.047 ± 0.001 \\
 & +M & 0.033 ± 0.006 & 0.020 ± 0.002 & 0.031 ± 0.000 & 0.047 ± 0.001 \\
 & +CT & 0.035 ± 0.007 & 0.027 ± 0.004 & 0.024 ± 0.000 & 0.064 ± 0.001 \\
\cline{1-6}
\bottomrule
\end{tabular}
\end{table}

\begin{table}[H]
\caption{Performance across feature sets and models for Number of Children (overall split). Values are mean±sd across folds; best per model in \textbf{bold}; best in metric in \textcolor{red}{\textbf{red}}.}
\label{tab:child_overall}
\begin{tabular}{llcccc}
\toprule
 &  & DNN & RF & GBM & SVM \\
Metric & Feature set &  &  &  &  \\
\midrule
\multirow{5}{*}{\rotatebox[origin=c]{90}{Accuracy}} & C & 0.776 ± 0.001 & 0.783 ± 0.001 & 0.779 ± 0.000 & 0.774 ± 0.000 \\
 & +ST & 0.780 ± 0.002 & 0.807 ± 0.001 & 0.784 ± 0.000 & 0.782 ± 0.000 \\
 & +D & 0.782 ± 0.001 & 0.805 ± 0.000 & 0.783 ± 0.000 & 0.786 ± 0.000 \\
 & +M & 0.781 ± 0.002 & 0.804 ± 0.001 & 0.781 ± 0.000 & 0.786 ± 0.001 \\
 & +CT & \textbf{0.785 ± 0.002} & \textcolor{red}{\textbf{0.808 ± 0.001}} & \textbf{0.787 ± 0.000} & \textbf{0.791 ± 0.000} \\
\cline{1-6}
\multirow{5}{*}{\rotatebox[origin=c]{90}{AUROC}} & C & 0.837 ± 0.001 & 0.823 ± 0.001 & 0.842 ± 0.000 & 0.831 ± 0.000 \\
 & +ST & 0.847 ± 0.002 & 0.861 ± 0.001 & 0.852 ± 0.000 & 0.844 ± 0.000 \\
 & +D & 0.849 ± 0.001 & 0.862 ± 0.002 & 0.853 ± 0.000 & 0.845 ± 0.000 \\
 & +M & 0.847 ± 0.003 & 0.861 ± 0.001 & 0.854 ± 0.000 & 0.844 ± 0.000 \\
 & +CT & \textbf{0.854 ± 0.002} & \textcolor{red}{\textbf{0.862 ± 0.001}} & \textbf{0.859 ± 0.000} & \textbf{0.851 ± 0.000} \\
\cline{1-6}
\multirow{5}{*}{\rotatebox[origin=c]{90}{NLL}} & C & 0.639 ± 0.001 & 0.801 ± 0.008 & 0.634 ± 0.000 & 0.650 ± 0.000 \\
 & +ST & 0.620 ± 0.003 & 0.595 ± 0.003 & 0.615 ± 0.000 & 0.625 ± 0.000 \\
 & +D & 0.619 ± 0.002 & 0.597 ± 0.008 & 0.613 ± 0.000 & 0.623 ± 0.000 \\
 & +M & 0.621 ± 0.004 & 0.596 ± 0.005 & 0.613 ± 0.000 & 0.623 ± 0.000 \\
 & +CT & \textbf{0.610 ± 0.003} & \textcolor{red}{\textbf{0.595 ± 0.010}} & \textbf{0.604 ± 0.000} & \textbf{0.613 ± 0.000} \\
\cline{1-6}
\multirow{5}{*}{\rotatebox[origin=c]{90}{ECE}} & C & 0.018 ± 0.001 & \textbf{0.030 ± 0.003} & 0.018 ± 0.000 & 0.059 ± 0.001 \\
 & +ST & 0.021 ± 0.003 & 0.055 ± 0.001 & 0.019 ± 0.000 & 0.045 ± 0.002 \\
 & +D & 0.022 ± 0.003 & 0.057 ± 0.002 & 0.023 ± 0.000 & 0.035 ± 0.001 \\
 & +M & 0.023 ± 0.004 & 0.058 ± 0.001 & 0.020 ± 0.000 & 0.038 ± 0.001 \\
 & +CT & \textcolor{red}{\textbf{0.016 ± 0.006}} & 0.048 ± 0.002 & \textbf{0.017 ± 0.000} & \textbf{0.030 ± 0.001} \\
\cline{1-6}
\bottomrule
\end{tabular}
\end{table}

\subsection{Cross-temporal Split}

\begin{table}[H]
\caption{Performance across feature sets and models for Age (2017–2019 train, 2023 test split). Values are mean±sd across folds; best per model in \textbf{bold}; best in metric in \textcolor{red}{\textbf{red}}.}
\label{tab:age_2017_2019_train_2023_test}
\begin{tabular}{llcccc}
\toprule
 &  & DNN & RF & GBM & SVM \\
Metric & Feature set &  &  &  &  \\
\midrule
\multirow{5}{*}{\rotatebox[origin=c]{90}{Accuracy}} & C & 0.474 ± 0.002 & 0.443 ± 0.001 & 0.471 ± 0.000 & 0.469 ± 0.000 \\
 & +ST & \textbf{0.488 ± 0.004} & 0.480 ± 0.002 & 0.489 ± 0.000 & 0.484 ± 0.001 \\
 & +D & 0.480 ± 0.003 & 0.479 ± 0.004 & 0.485 ± 0.000 & 0.484 ± 0.001 \\
 & +M & 0.475 ± 0.004 & 0.478 ± 0.003 & 0.483 ± 0.000 & 0.480 ± 0.001 \\
 & +CT & 0.484 ± 0.004 & \textbf{0.482 ± 0.002} & \textcolor{red}{\textbf{0.490 ± 0.001}} & \textbf{0.485 ± 0.000} \\
\cline{1-6}
\multirow{5}{*}{\rotatebox[origin=c]{90}{AUROC}} & C & 0.814 ± 0.001 & 0.781 ± 0.001 & 0.813 ± 0.000 & 0.802 ± 0.000 \\
 & +ST & \textbf{0.826 ± 0.001} & 0.813 ± 0.000 & 0.824 ± 0.000 & \textbf{0.815 ± 0.000} \\
 & +D & 0.824 ± 0.001 & 0.812 ± 0.001 & 0.824 ± 0.000 & 0.814 ± 0.000 \\
 & +M & 0.821 ± 0.001 & 0.812 ± 0.001 & 0.823 ± 0.000 & 0.813 ± 0.000 \\
 & +CT & 0.824 ± 0.001 & \textbf{0.815 ± 0.001} & \textcolor{red}{\textbf{0.826 ± 0.000}} & 0.815 ± 0.000 \\
\cline{1-6}
\multirow{5}{*}{\rotatebox[origin=c]{90}{NLL}} & C & 1.181 ± 0.004 & 1.505 ± 0.006 & 1.176 ± 0.000 & 1.216 ± 0.001 \\
 & +ST & \textbf{1.148 ± 0.005} & 1.186 ± 0.001 & 1.147 ± 0.000 & 1.184 ± 0.001 \\
 & +D & 1.158 ± 0.004 & 1.190 ± 0.005 & 1.149 ± 0.000 & 1.188 ± 0.001 \\
 & +M & 1.164 ± 0.006 & 1.189 ± 0.002 & 1.151 ± 0.000 & 1.193 ± 0.001 \\
 & +CT & 1.153 ± 0.006 & \textbf{1.173 ± 0.004} & \textcolor{red}{\textbf{1.135 ± 0.000}} & \textbf{1.180 ± 0.000} \\
\cline{1-6}
\multirow{5}{*}{\rotatebox[origin=c]{90}{ECE}} & C & \textbf{0.034 ± 0.003} & 0.087 ± 0.001 & 0.039 ± 0.000 & 0.055 ± 0.001 \\
 & +ST & 0.037 ± 0.007 & \textcolor{red}{\textbf{0.015 ± 0.003}} & 0.034 ± 0.000 & 0.051 ± 0.002 \\
 & +D & 0.044 ± 0.006 & 0.016 ± 0.002 & \textbf{0.034 ± 0.000} & \textbf{0.051 ± 0.001} \\
 & +M & 0.047 ± 0.004 & 0.017 ± 0.006 & 0.039 ± 0.000 & 0.055 ± 0.001 \\
 & +CT & 0.050 ± 0.005 & 0.019 ± 0.003 & 0.036 ± 0.001 & 0.057 ± 0.001 \\
\cline{1-6}
\bottomrule
\end{tabular}
\end{table}

\begin{table}[H]
\caption{Performance across feature sets and models for Gender (2017–2019 train, 2023 test split). Values are mean±sd across folds; best per model in \textbf{bold}; best in metric in \textcolor{red}{\textbf{red}}.}
\label{tab:gender_2017_2019_train_2023_test}
\begin{tabular}{llcccc}
\toprule
 &  & DNN & RF & GBM & SVM \\
Metric & Feature set &  &  &  &  \\
\midrule
\multirow{5}{*}{\rotatebox[origin=c]{90}{Accuracy}} & C & 0.507 ± 0.003 & 0.500 ± 0.003 & 0.506 ± 0.000 & 0.508 ± 0.000 \\
 & +ST & 0.508 ± 0.003 & \textbf{0.517 ± 0.002} & \textcolor{red}{\textbf{0.524 ± 0.000}} & 0.511 ± 0.001 \\
 & +D & 0.511 ± 0.002 & 0.513 ± 0.002 & 0.523 ± 0.000 & 0.513 ± 0.000 \\
 & +M & 0.507 ± 0.002 & 0.509 ± 0.003 & 0.523 ± 0.000 & 0.509 ± 0.001 \\
 & +CT & \textbf{0.512 ± 0.006} & 0.515 ± 0.002 & 0.523 ± 0.000 & \textbf{0.514 ± 0.000} \\
\cline{1-6}
\multirow{5}{*}{\rotatebox[origin=c]{90}{AUROC}} & C & 0.551 ± 0.001 & 0.514 ± 0.001 & 0.546 ± 0.000 & 0.542 ± 0.000 \\
 & +ST & \textbf{0.557 ± 0.003} & 0.544 ± 0.003 & 0.567 ± 0.000 & \textbf{0.547 ± 0.000} \\
 & +D & 0.556 ± 0.003 & 0.540 ± 0.003 & 0.565 ± 0.000 & 0.540 ± 0.000 \\
 & +M & 0.554 ± 0.003 & 0.540 ± 0.001 & 0.568 ± 0.000 & 0.540 ± 0.000 \\
 & +CT & 0.556 ± 0.001 & \textbf{0.544 ± 0.002} & \textcolor{red}{\textbf{0.570 ± 0.000}} & 0.545 ± 0.000 \\
\cline{1-6}
\multirow{5}{*}{\rotatebox[origin=c]{90}{NLL}} & C & 0.961 ± 0.003 & 1.396 ± 0.008 & 0.976 ± 0.000 & \textcolor{red}{\textbf{0.953 ± 0.001}} \\
 & +ST & 0.962 ± 0.003 & 1.002 ± 0.005 & 0.969 ± 0.000 & 0.954 ± 0.000 \\
 & +D & 0.960 ± 0.006 & 1.002 ± 0.007 & 0.964 ± 0.000 & 0.954 ± 0.000 \\
 & +M & \textbf{0.960 ± 0.004} & \textbf{1.001 ± 0.006} & \textbf{0.961 ± 0.000} & 0.954 ± 0.000 \\
 & +CT & 0.963 ± 0.004 & 1.007 ± 0.011 & 0.962 ± 0.000 & 0.954 ± 0.001 \\
\cline{1-6}
\multirow{5}{*}{\rotatebox[origin=c]{90}{ECE}} & C & 0.056 ± 0.005 & 0.108 ± 0.003 & 0.058 ± 0.000 & 0.039 ± 0.002 \\
 & +ST & 0.060 ± 0.004 & \textbf{0.057 ± 0.003} & 0.046 ± 0.000 & 0.041 ± 0.002 \\
 & +D & \textbf{0.053 ± 0.003} & 0.059 ± 0.002 & 0.045 ± 0.000 & \textcolor{red}{\textbf{0.034 ± 0.001}} \\
 & +M & 0.060 ± 0.006 & 0.062 ± 0.003 & 0.044 ± 0.000 & 0.039 ± 0.001 \\
 & +CT & 0.060 ± 0.004 & 0.058 ± 0.002 & \textbf{0.044 ± 0.000} & 0.037 ± 0.002 \\
\cline{1-6}
\bottomrule
\end{tabular}
\end{table}

\begin{table}[H]
\caption{Performance across feature sets and models for HH Income (2017–2019 train, 2023 test split). Values are mean±sd across folds; best per model in \textbf{bold}; best in metric in \textcolor{red}{\textbf{red}}.}
\label{tab:income_2017_2019_train_2023_test}
\begin{tabular}{llcccc}
\toprule
 &  & DNN & RF & GBM & SVM \\
Metric & Feature set &  &  &  &  \\
\midrule
\multirow{5}{*}{\rotatebox[origin=c]{90}{Accuracy}}
 & C   & 0.588 ± 0.001 & 0.542 ± 0.002 & \textbf{0.582 ± 0.000} & \textbf{0.586 ± 0.001} \\
 & +ST & 0.588 ± 0.002 & \textbf{0.569 ± 0.002} & 0.579 ± 0.000 & 0.586 ± 0.001 \\
 & +D  & 0.587 ± 0.002 & 0.566 ± 0.002 & 0.579 ± 0.000 & 0.586 ± 0.001 \\
 & +M  & \textcolor{red}{\textbf{0.588 ± 0.001}} & 0.566 ± 0.001 & 0.581 ± 0.000 & 0.585 ± 0.001 \\
 & +CT & 0.585 ± 0.002 & 0.562 ± 0.002 & 0.577 ± 0.000 & 0.581 ± 0.001 \\
\cline{1-6}
\multirow{5}{*}{\rotatebox[origin=c]{90}{AUROC}}
 & C   & 0.606 ± 0.003 & 0.581 ± 0.000 & 0.605 ± 0.000 & 0.583 ± 0.000 \\
 & +ST & 0.616 ± 0.004 & 0.610 ± 0.002 & 0.617 ± 0.000 & 0.599 ± 0.000 \\
 & +D  & 0.614 ± 0.002 & 0.610 ± 0.001 & \textcolor{red}{\textbf{0.620 ± 0.000}} & 0.598 ± 0.000 \\
 & +M  & 0.617 ± 0.002 & 0.610 ± 0.002 & 0.615 ± 0.000 & \textbf{0.604 ± 0.000} \\
 & +CT & \textbf{0.618 ± 0.003} & \textbf{0.611 ± 0.002} & 0.613 ± 0.000 & 0.604 ± 0.000 \\
\cline{1-6}
\multirow{5}{*}{\rotatebox[origin=c]{90}{NLL}}
 & C   & 1.219 ± 0.003 & 1.557 ± 0.022 & 1.228 ± 0.000 & 1.230 ± 0.000 \\
 & +ST & \textcolor{red}{\textbf{1.213 ± 0.005}} & \textbf{1.244 ± 0.002} & 1.225 ± 0.000 & 1.224 ± 0.000 \\
 & +D  & 1.214 ± 0.005 & 1.244 ± 0.002 & \textbf{1.221 ± 0.000} & 1.225 ± 0.000 \\
 & +M  & 1.214 ± 0.003 & 1.244 ± 0.002 & 1.226 ± 0.000 & 1.222 ± 0.000 \\
 & +CT & 1.219 ± 0.005 & 1.244 ± 0.002 & 1.225 ± 0.000 & \textbf{1.220 ± 0.000} \\
\cline{1-6}
\multirow{5}{*}{\rotatebox[origin=c]{90}{ECE}}
 & C   & 0.063 ± 0.004 & 0.065 ± 0.004 & 0.067 ± 0.000 & 0.069 ± 0.001 \\
 & +ST & 0.058 ± 0.002 & 0.080 ± 0.001 & 0.068 ± 0.000 & 0.070 ± 0.001 \\
 & +D  & 0.052 ± 0.008 & 0.079 ± 0.001 & 0.064 ± 0.000 & 0.072 ± 0.001 \\
 & +M  & 0.057 ± 0.006 & 0.080 ± 0.002 & 0.070 ± 0.000 & 0.072 ± 0.001 \\
 & +CT & \textbf{0.047 ± 0.008} & \textbf{0.054 ± 0.002} & \textcolor{red}{\textbf{0.043 ± 0.000}} & \textbf{0.058 ± 0.001} \\
\cline{1-6}
\bottomrule
\end{tabular}
\end{table}

\begin{table}[H]
\caption{Performance across feature sets and models for Number of Children (2017–2019 train, 2023 test split). Values are mean±sd across folds; best per model in \textbf{bold}; best in metric in \textcolor{red}{\textbf{red}}.}
\label{tab:child_2017_2019_train_2023_test}
\begin{tabular}{llcccc}
\toprule
 &  & DNN & RF & GBM & SVM \\
Metric & Feature set &  &  &  &  \\
\midrule
\multirow{5}{*}{\rotatebox[origin=c]{90}{Accuracy}}
 & C   & 0.747 ± 0.002 & 0.741 ± 0.003 & 0.745 ± 0.000 & 0.754 ± 0.000 \\
 & +ST & \textbf{0.756 ± 0.003} & \textbf{0.759 ± 0.001} & 0.749 ± 0.000 & 0.758 ± 0.001 \\
 & +D  & 0.751 ± 0.001 & 0.758 ± 0.002 & 0.747 ± 0.000 & \textcolor{red}{\textbf{0.760 ± 0.000}} \\
 & +M  & 0.752 ± 0.003 & 0.757 ± 0.001 & \textbf{0.750 ± 0.000} & 0.756 ± 0.001 \\
 & +CT & 0.753 ± 0.002 & 0.758 ± 0.001 & 0.747 ± 0.000 & 0.757 ± 0.000 \\
\cline{1-6}
\multirow{5}{*}{\rotatebox[origin=c]{90}{AUROC}}
 & C   & 0.819 ± 0.002 & 0.791 ± 0.001 & 0.819 ± 0.000 & 0.808 ± 0.000 \\
 & +ST & 0.825 ± 0.002 & 0.809 ± 0.001 & 0.825 ± 0.000 & 0.814 ± 0.000 \\
 & +D  & 0.820 ± 0.001 & 0.811 ± 0.002 & 0.826 ± 0.000 & 0.814 ± 0.000 \\
 & +M  & 0.821 ± 0.001 & 0.810 ± 0.001 & 0.826 ± 0.000 & 0.814 ± 0.000 \\
 & +CT & \textbf{0.826 ± 0.001} & \textbf{0.816 ± 0.001} & \textcolor{red}{\textbf{0.828 ± 0.000}} & \textbf{0.817 ± 0.000} \\
\cline{1-6}
\multirow{5}{*}{\rotatebox[origin=c]{90}{NLL}}
 & C   & 0.704 ± 0.003 & 1.031 ± 0.016 & 0.712 ± 0.000 & 0.717 ± 0.000 \\
 & +ST & \textcolor{red}{\textbf{0.690 ± 0.002}} & 0.729 ± 0.005 & 0.695 ± 0.000 & 0.705 ± 0.000 \\
 & +D  & 0.695 ± 0.002 & 0.722 ± 0.006 & 0.695 ± 0.000 & \textbf{0.704 ± 0.000} \\
 & +M  & 0.696 ± 0.005 & \textbf{0.720 ± 0.005} & 0.695 ± 0.000 & 0.706 ± 0.001 \\
 & +CT & 0.696 ± 0.004 & 0.726 ± 0.004 & \textbf{0.692 ± 0.000} & 0.707 ± 0.001 \\
\cline{1-6}
\multirow{5}{*}{\rotatebox[origin=c]{90}{ECE}}
 & C   & 0.025 ± 0.002 & \textbf{0.038 ± 0.004} & 0.020 ± 0.000 & 0.034 ± 0.001 \\
 & +ST & \textbf{0.021 ± 0.004} & 0.040 ± 0.001 & 0.016 ± 0.000 & 0.023 ± 0.001 \\
 & +D  & 0.024 ± 0.003 & 0.040 ± 0.002 & 0.016 ± 0.000 & 0.022 ± 0.001 \\
 & +M  & 0.026 ± 0.003 & 0.041 ± 0.001 & \textcolor{red}{\textbf{0.013 ± 0.000}} & \textbf{0.017 ± 0.001} \\
 & +CT & 0.026 ± 0.002 & 0.039 ± 0.001 & 0.015 ± 0.001 & 0.019 ± 0.001 \\
\cline{1-6}
\bottomrule
\end{tabular}
\end{table}

\end{appendix}





\end{document}